\pdfoutput=1

\documentclass[11pt]{article}

\usepackage{ACL2023}

\usepackage{times}
\usepackage{latexsym}

\usepackage[T1]{fontenc}

\usepackage[utf8]{inputenc}

\usepackage{microtype}

\usepackage{fontawesome5}
\usepackage{booktabs}
\usepackage{graphicx}
\usepackage{multirow}
\usepackage{subcaption}
\usepackage{color, colortbl}
\usepackage{graphics}
\usepackage{amsmath}
\usepackage{lipsum}
\usepackage{algorithm}
\usepackage{algpseudocode}
\usepackage{comment}
\usepackage{makecell}
\usepackage{bm}
\usepackage{amssymb}
\usepackage[most]{tcolorbox}

\def\argmax{\operatornamewithlimits{arg\,max}}

\usepackage[colorinlistoftodos,prependcaption,textsize=tiny]{todonotes}

\usepackage{inconsolata}

\title{FLAP: Flow-Adhering Planning with Constrained Decoding in LLMs}

\author{
Shamik Roy \quad Sailik Sengupta \quad Daniele Bonadiman \quad Saab Mansour \quad Arshit Gupta\\
  \texttt{\{royshami, sailiks, dbonadim, saabm, arshig\}@amazon.com}\\  
  \faAmazon WS AI Labs
  }

\begin{document}
\maketitle
\begin{abstract}
Planning is a crucial task for agents in task oriented dialogs (TODs). Human agents typically resolve user issues by following predefined workflows, decomposing workflow steps into actionable items, and performing actions by executing APIs in order; all of which require reasoning and planning. With the recent advances in LLMs, there have been increasing attempts to use them for task planning and API usage. However, the faithfulness of the plans to predefined workflows and API dependencies, is not guaranteed with LLMs. 
Moreover, workflows in real life are often custom-defined and prone to changes; hence, adaptation is desirable. To study this, we propose the problem of faithful planning in TODs that needs to resolve user intents by following predefined flows and preserving API dependencies. To solve this problem, we propose \textbf{FLAP}, a \textbf{Fl}ow-\textbf{A}dhering \textbf{P}lanning algorithm based on constrained decoding with lookahead heuristic for LLMs. Our algorithm alleviates the need for finetuning LLMs using domain specific (plan/dependency) data, enables quick adaptation to predefined flows, and outperforms other decoding and prompting-based baselines. Further, our algorithm empowers smaller LLMs ($\approx7$B) to perform at par larger LLMs ($\approx30$B-$40$B).
\end{abstract}

\section{Introduction}
\label{sec:introduction}
\begin{figure}[t!]
  \includegraphics[width=0.48\textwidth]{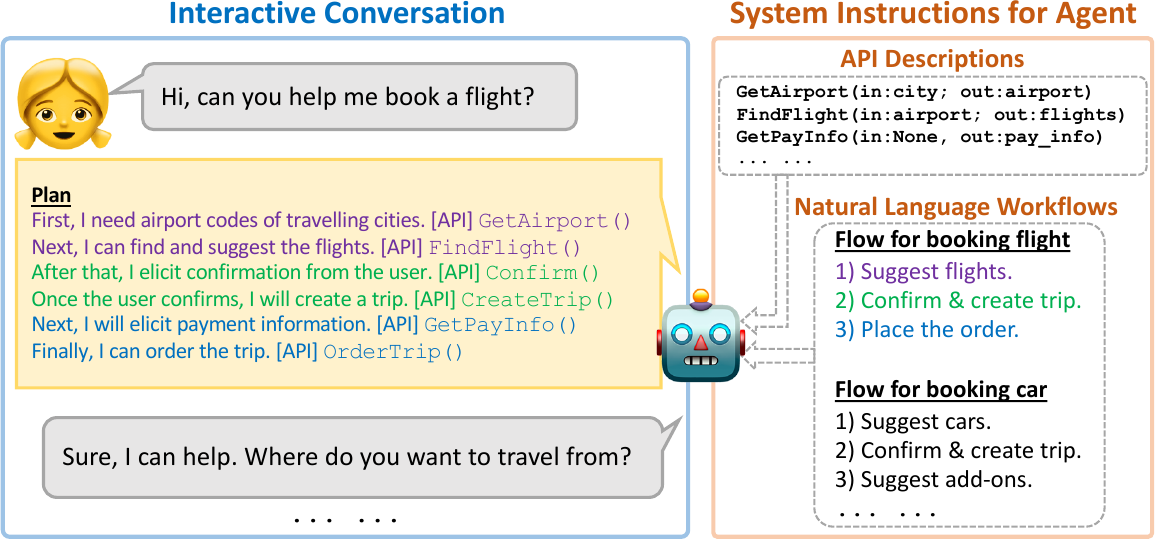}
  \caption{An agent-user interaction scenario. Agent plans how to resolve the user query by following given natural language workflows and API dependencies.} 
  \label{fig:faithfil-plan-generation-task-example}
\end{figure}

A successful task oriented dialog (TOD) often involves interaction with internal or external world through tools or APIs. For example, as shown in Fig. \ref{fig:faithfil-plan-generation-task-example}, booking a flight through a conversation may involve searching for flights, getting payment information, and so on. API usage is a difficult task for chatbot agents as it involves various reasoning and planning. Beyond predicting the correct API to use in a particular scenario, it requires task decomposition and resolution of inter-API dependencies (e.g., searching for flights requires extracting airport codes) \citep{zhang2021learning,prasad2023adapt,wang-etal-2023-measuring,lu2023gear}. Moreover, there may exist customized workflows to follow for API usage, such as ``confirm before placing order'' or ``recommend add-ons before checking out'', which can conveniently be defined using natural language (NL) instructions. %

Recent advances in LLMs have enabled researchers to consider LLMs for task-planning \cite{huang2022language,valmeekam2023on} and API usage \cite{qin2023toolllm,patil2023gorilla}. In planning, studies have primarily focused on open-ended scenarios and emphasized mostly on goal attainment \citep{shridhar2020alfred,jansen2020visually} without considering any instructions on the process \cite{valmeekam2024planbench}, while works on API usage have mostly considered the appropriateness of API calls \cite{wang-etal-2023-measuring}. However, considerations about API dependencies, and the faithfulness of generated API calls to predefined workflows or instructions are understudied.

In this paper, we propose a novel problem of zero-shot faithful planning with LLMs in TODs. As input, we provide (1) the domain definition in terms of APIs and their descriptions, (2) NL flows to fulfill various intents in the domain, and (3) a user query. The task is to generate a plan that fulfills the user's query by using the domain-specific APIs while adhering to the specified flows and API dependencies (\S\ref{sec:task-formulation}).

Prior works attempted to teach LLMs to use APIs by pretraining them with large amount of API related data \citep{schick2023toolformer,yang2023gpt4tools,qin2023toolllm}. However, in practice, API signatures and the NL flows to use them are customized to use cases and are prone to change (e.g., due to change in business policy). Hence, quickly adapting the LLMs to new flows and API dependencies is required. Unfortunately, pretraining LLMs cannot guarantee faithful planning with custom-defined APIs and flows which may significantly differ from the pretraining data \citep{blodgett-etal-2020-language,bommasani2021opportunities}; 
we experimentally verify this in \S\ref{sec:ablation}. To alleviate the issue, we propose \textbf{FLAP}, a \textbf{Fl}ow \textbf{A}dhering \textbf{P}lanning algorithm that uses constrained decoding of LLMs with lookahead heuristic \citep{och2001efficient,haghighi2007approximate,lu2022neurologic,huang2024deal} (\S\ref{sec:proposed-algorithm}). FLAP generates faithful plans to predefined flows and API dependencies, requires no pretraining, and is applicable on top of any autoregressive LLM regardless of its pretraining mechanism.

Given previous studies have shown that thoughts enhance the reasoning capabilities of LLMs \cite{wei2022chain} (even in the context of API usage \cite{yao2022react}), FLAP prompts the LLMs to generate plans consisting of thoughts and corresponding APIs. FLAP aligns the generated thoughts to predefined NL flows, enabling thoughts to act as an intermediary (and address the vocabulary mismatch) between NL flows and APIs. In addition, FLAP captures the relations among NL flow steps and APIs using dynamic dependency graphs; these are used to score different dimensions related to API and flow dependency adherence in the lookahead heuristic function. Finally, FLAP also enforces structural constraints using lookahead heuristics.

To quantify the efficacy of current LLMs with promising prompt strategies, and our proposed approach for {\em faithful planning}, we construct a novel dataset comprising $4$ domains, $13$ intents and flows, $64$ APIs, and $260$ user queries, resembling real life use-cases (\S\ref{sec:data-collection}). We find that existing open-source LLMs and prompting techniques lack faithful planning capabilities and our proposed constrained decoding based algorithm, FLAP, improves the faithfulness of plans by reducing various errors, such as API and flow step dependency violations, redundant/lacking API in generated plan, etc. We also demonstrate that smaller LLMs ($7$B) with FLAP can achieve comparable performance against much larger LLMs ($40$B) (\S\ref{sec:experiments}).

\section{Task Formulation}
\label{sec:task-formulation}
In this section, we define the novel task of faithful planning in task oriented dialogues (TODs) to resolve queries by adhering to predefined natural language (NL) flows and API dependencies. 

\subsection{TOD Agent Scenario}
In a typical TOD system, an agent is equipped with the following elements. 

\noindent\textbf{Domain:} An agent usually has access to a specific domain, $\mathcal{D}$ (e.g., trip planning, banking, and so on).

\noindent\textbf{Intents:} In a domain $\mathcal{D}$, an agent has capabilities to handle a set of intents, $\mathcal{I}$. For example, an agent form the domain ``trip planner'' may only be able to book a hotel, search flights, and so on, however, unlikely to be able to create a bank account.

\noindent\textbf{Flows:} Agents are often provided with a sequence of natural language instructions, defined as ``flows'', $\mathcal{F}$ (e.g., ``recommend add-ons before checking out''). Agents are expected to follow these flows \textit{step-by-step} to resolve intents. Flows are typically defined by businesses and are prone to change.

\noindent\textbf{APIs:} Agents have access to a set of internal APIs, $\mathcal{A}$, to interact with the system (typically through GUI) and trigger actions based on the inputs from the user. For example, if a user is looking to book a car in Chicago, the agent may trigger the API, \texttt{FindRentalCar(location=``Chicago'')}.

\subsection{Faithful Planning for Resolving Intent}
In real life, when a user expresses an intent, an agent has to execute a sequence of APIs to fulfil it. Successful fulfillment of the user intent depends largely on the reasoning and planning capability of the agents because of the following challenges.

\noindent\textbf{Identifying and following NL flows:} Given an intent in $\mathcal{I}$, the execution steps need to be strictly consistent with a predefined flow in $\mathcal{F}$. Hence, the agents need to first identify the correct flow in $\mathcal{F}$ and then adhere to it.

\noindent\textbf{Task decomposition and dependency resolution:} APIs in $\mathcal{A}$  may have dependencies on other APIs in $\mathcal{A}$, and a flow step may require calling multiple APIs. For example, \texttt{GetAirports(cit-} \texttt{y=``Chicago'')} is needed to be called before \texttt{Find-} \texttt{Flight(airport=``ORD'')}, to execute the flow step ``suggest flights''. Hence, the agents have to decompose flow steps and preserve API dependencies.

Grounded in such challenges, we define the novel task of faithful planning for TOD agents (Fig. \ref{fig:faithfil-plan-generation-task-example}). Given, a domain $\mathcal{D}$, a set of flows $\mathcal{F}$ (for resolving intents $\mathcal{I}$), a set of APIs $\mathcal{A}$ (with parameters and descriptions), and a user query $\mathcal{Q}$; the task is to generate a plan $\mathcal{P}$ consisting of a sequence of APIs to fulfil the user query while obeying the flow steps in $\mathcal{F}$ and preserving API dependencies in $\mathcal{A}$. 

In real world scenarios, agents may need to dynamically update the plan based on the situation (say, the user starts with query $\mathcal{Q}$ and then wants to do $\mathcal{Q}^\prime$ as well). As a first step towards faithful planning in TODs, we study static planning (user does not change mind) in this paper, and leave dynamic (run-time) planning as a future work.

\subsection{Zero-shot Planning with LLMs}
\label{sec:zero-shot-planning-with-llms}
We study the capabilities of LLMs in zero-shot planning by following predefined flows and preserving API dependencies in TODs. For that, we instruct a base LLM to act as a customer care agent and in context, we provide the domain definition consisting of APIs, their input-output parameters and descriptions, and NL flows for resolving intents in the domain. Given a user query in context, the LLM is instructed to plan how to resolve the query by using the given APIs and following the appropriate flow. The LLMs are expected to resolve API dependencies from their input-output parameters. The prompt structures are shown in \S\ref{sec:appx-prompt-structure-for-plan-generation}. We study the problem in the following settings by varying the nature and complexity of the task.

\noindent\textbf{All \textit{vs.} relevant flow in-context:} We experiment by providing either all flows or only the flow related to the user query in context. The former setting is more complex as the LLM has to first identify the relevant flow and then follow it to plan.

\noindent\textbf{Reasoning and acting:} Recent studies have shown that step-by-step thinking and acting improves the end goal achievement \citep{wei2022chain,yao2022react}. Inspired by these works, we study the problem of plan generation in a setting where the LLMs are prompted to generate a thought and then an API in the plan to help it rationalize its selection of the API.

\subsection{Evaluation Metrics}

\noindent\textbf{Number of edits to fix the plan:} A generated plan is required to contain \textit{only} the steps and APIs from the gold plan. We define number of edits as the sum of additions and deletions of steps/APIs in the generated plan to make it contain the same steps/APIs as the gold plan. The lower the number of edits, the better the generated plan.
    
\noindent\textbf{Inconsistent steps/APIs:} A generated plan is required to obey flow steps and API dependencies. Hence, we measure the percentage of steps or APIs in the generated plan that do not follow the dependency structure (e.g., a step/API is generated before its parent steps/APIs). 

\noindent\textbf{Fine-grained metrics:} We quantify other fine-grained metrics such as API hallucination, API repetition and parsability of the generated plans.

\begin{table}[t!]
\centering
\begin{subtable}{1\columnwidth}\centering
\resizebox{1\columnwidth}{!}{%
\begin{tabular}
{>{\arraybackslash}m{4.6cm}>{\centering\arraybackslash}m{1.5cm}|>{\centering\arraybackslash}m{1.5cm}|>{\centering\arraybackslash}m{1.5cm}|>{\centering\arraybackslash}m{2.1cm}}
\toprule

\textbf{Domains:} & \textbf{Trip Booking} & \textbf{Insurance} & \textbf{Banking} & \textbf{Restaurant \& Ride Book}\\
\midrule
\# of intents/flows             & $3$ & $3$ & $3$ & $4$ \\
\# of steps per intent/flow     & $5$-$6$ & $4$-$5$ & $3$-$5$ & $4$\\
\# of APIs needed per step      & $1$-$2$ & $1$-$2$ & $1$-$3$ & $1$-$4$\\
\# of APIs in domain            & $13$ & $15$ & $14$ & $22$\\
\# of relationships among APIs  & $13$ & $13$ & $15$ & $19$\\
\# of user queries           & $60$  & $60$ & $60$ & $80$\\
\end{tabular}}

\centering
\resizebox{1\columnwidth}{!}{%
\begin{tabular}
{>{\arraybackslash}m{3cm}|>{\arraybackslash}m{4cm}|>{\arraybackslash}m{4cm}|>{\arraybackslash}m{4.5cm}}
\toprule

\multicolumn{4}{c}{\textbf{Intents in all domains}}\\
\textbf{Trip Booking} & \textbf{Insurance} & \textbf{Banking} & \textbf{Restaurant \& Ride Book}\\
book car, book hotel, book flight & buy insurance, cancel insurance, add member & open account, report problem, cancel transaction & book restaurant, book ride, cancel restaurant, cancel ride\\
\bottomrule
\end{tabular}}
\centering
\caption{Created domains statistics.}
\label{tab:domain-and-flow-statistics}
\end{subtable}

\begin{subtable}{1\columnwidth}\centering
\resizebox{1\columnwidth}{!}{%
\begin{tabular}
{>{\arraybackslash}m{5.2cm}|>{\arraybackslash}m{7.5cm}}
\midrule
\textbf{Flow steps} & \textbf{Required API sequence (gold plan)}\\
\midrule
Suggest cars to the customer                    & \texttt{FindRentalCar()}\\
Confirm and create the trip                     & \texttt{Confirm(), CreateTrip()}\\
Extract and add promotional offers              & \texttt{GetCarInsuranceDiscount(), UpdateTrip()}\\
Order the trip                                  & \texttt{GetPaymentInformation(), OrderTrip()}\\
\bottomrule
\end{tabular}}
\caption{Flow and required APIs for the intent ``book car'' in the domain ``Trip Booking''. The full list can be found in \S\ref{sec:appx-domain-creation}.}
\label{tab:domain-and-flow-example}
\end{subtable}

\caption{Test data statistics (a) and example flow (b).}
\label{tab:domain-dataset-summary}
\end{table}

\begin{table*}[ht!]
\centering
\resizebox{2\columnwidth}{!}{%
\begin{tabular}
{>{\arraybackslash}m{3cm}>{\centering\arraybackslash}m{1.3cm}|>{\centering\arraybackslash}m{1.75cm}|>{\centering\arraybackslash}m{1.4cm}>{\centering\arraybackslash}m{1.35cm}|>{\centering\arraybackslash}m{1.5cm}>{\centering\arraybackslash}m{1.5cm}||>{\centering\arraybackslash}m{1.75cm}|>{\centering\arraybackslash}m{1.4cm}>{\centering\arraybackslash}m{1.35cm}|>{\centering\arraybackslash}m{1.5cm}>{\centering\arraybackslash}m{1.5cm}}
\toprule

&  & \multicolumn{5}{c||}{\textbf{All flows in-context}} & \multicolumn{5}{c}{\textbf{Only relevant flow in-context}}\\

\cmidrule{3-12}

& \textbf{With} & \textbf{Avg \% of} & \multicolumn{2}{c|}{\textbf{Avg \# of edit for}}	& \multicolumn{2}{c||}{\textbf{Avg \% of inconsistent}} & \textbf{Avg \% of} & \multicolumn{2}{c|}{\textbf{Avg \# of edit for}}	& \multicolumn{2}{c}{\textbf{Avg \% of inconsistent}}	\\

\textbf{LLMs} & \textbf{Thought} & \textbf{API Hallu.} & \textbf{Steps ($\downarrow$)}	& \textbf{APIs ($\downarrow$)}	& \textbf{Steps ($\downarrow$)}	& \textbf{APIs ($\downarrow$)} & \textbf{API Hallu.} & \textbf{Steps ($\downarrow$)}	& \textbf{APIs ($\downarrow$)}	& \textbf{Steps ($\downarrow$)}	& \textbf{APIs ($\downarrow$)}\\
\toprule
\multirow{2}{*}{\textbf{\texttt{santacoder-1.1b}}}  & No & $1.2\%$	& $30.3$	& $25.9$	& $15.4\%$	& $33.1\%$ & $1.8\%$ &  $29.6$ &  $25.8$ &  $15.3\%$ & $34.1\%$\\
                                                    & Yes & $13.0\%$	& $11.8$	& $13.0$	& $18.2\%$	& $34.1\%$ & $20.0\%$ &  $9.5$ &  $11.5$ &  $19.5\%$ & $31.2\%$\\
\midrule
\multirow{2}{*}{\textbf{\texttt{toolAlpaca-7b}}}    & No & $10.8\%$	&	$4.6$ &	$5.1$ &	$20.4\%$ &	$49.6\%$ & $7.0\%$ &	$4.4$ &	$4.7$ &	$17.8\%$ &	$49.2\%$\\
                                                    & Yes & $19.0\%$	&	$4.7$ &	$5.3$ &	$17.6\%$ &	$44.8\%$ & $8.0\%$ &	$3.0$ &	$3.5$ &	$10.6\%$ &	$47.6\%$\\                     
\midrule
\multirow{2}{*}{\textbf{\texttt{falcon-7b-inst.}}}	& No & $6.1\%$	& $47.5$	& $40.3$	& $16.6\%$	& $27.7\%$ & $6.6\%$ &  $40.8$ &  $38.1$ &  $17.4\%$ & $23.9\%$\\
                                                        & Yes & $19.0\%$	& $9.7$	& $11.4$	& $24.0\%$	& $38.8\%$ & $9.0\%$ &  $9.7$ &  $9.2$ &  $21.5\%$ & $37.9\%$\\
\midrule
\multirow{2}{*}{\textbf{\texttt{mpt-7b-inst.}}}  & No & $0.2\%$	& $16.2$	& $13.4$	& $14.7\%$	& $33.3\%$ & $0.2\%$ &  $13.6$ &  $11.5$ &  $13.5\%$ & $35.6\%$\\
                                                    & Yes & $4.0\%$	& $13.9$ &	$12.4$ &	$11.9\%$ &	$29.6\%$ & $3.0\%$ &  $9.9$ &	$8.9$ &	$9.1\%$ &	$33.4\%$\\
\midrule
\multirow{2}{*}{\textbf{\texttt{mistral-7b-inst.}}}  & No & $2.4\%$	& $2.9$	& $3.0$	& $10.6\%$	& $37.2\%$ & $1.4\%$ &  $3.1$ &  $3.1$ &  $8.0\%$ & $30.6\%$\\
                                                        & Yes & $0\%$	& $2.8$ &	$2.8$ &	$3.1\%$ &	$40.3\%$ & $1.0\%$ &  $2.7$ &	$2.6$ &	$2.8\%$ &	$34.4\%$\\
\midrule
\multirow{2}{*}{\textbf{\texttt{koala-13b}}}	& No & $6.3\%$	& $37.6$	& $31.4$	& $10.5\%$	& $24.0\%$ & $4.2\%$ &  $29.4$ &  $23.8$ &  $11.3\%$ & $26.2\%$\\
                                                & Yes & $10.0\%$	& $8.8$	& $9.1$	& $9.0\%$	& $24.0\%$ & $9.0\%$ & $7.4$ & $7.9$ & $9.3\%$ & $22.6\%$\\
\midrule
\multirow{2}{*}{\textbf{\texttt{vicuna-13b}}}   & No & $2.4\%$	& $4.1$	& $4.3$	& $15.1\%$	& $35.2\%$ & $2.4\%$ & $3.9$ & $4.1$ & $14.2\%$ & $31.7\%$\\
                                                & Yes & $4.0\%$	& $4.0$	& $4.3$	& $6.7\%$	& $35.5\%$ & $2.0\%$ & $3.1$ & $3.4$ & $5.4\%$ & $35.0\%$\\
\midrule
\multirow{2}{*}{\textbf{\texttt{llama-13b}}}    & No & $2.4\%$	& $26.8$	& $20.9$	& $13.2\%$	& $31.7\%$ & $2.0\%$ & $24.7$ & $19.1$ & $11.3\%$ & $29.4\%$\\
                                                & Yes & $4.0\%$	& $9.2$	& $8.7$	& $4.7\%$	& $31.0\%$ & $3.0\%$ & $6.6$ & $5.6$ & $4.0\%$ & $38.4\%$\\
\midrule
\multirow{2}{*}{\textbf{\texttt{mpt-30b-chat}}} & No & $0.5\%$	& $3.3$	& $3.4$	& $9.1\%$	& $33.4\%$ & $0.2\%$ & $3.5$ & $3.3$ & $6.3\%$ & $33.3\%$\\
                                                & Yes & $1.0\%$	& $2.7$	& $2.7$	& $5.7\%$	& $32.5\%$ & $1.0\%$ & $2.1$ & $2.1$ & $4.6\%$ & $34.3\%$\\
\midrule
\multirow{2}{*}{\textbf{\texttt{falcon-40b-inst.}}}  & No & $0.9\%$	& $8.1$	& $7.2$	& $18.0\%$	& $38.3\%$ & $0.9\%$ & $9.0$ & $7.9$ & $17.9\%$ & $35.5\%$\\
                                                        & Yes & $5.0\%$	& $5.4$	& $5.4$	& $9.0\%$	& $36.4\%$ & $7.0\%$ & $4.0$ & $4.1$ & $8.7\%$ & $25.7\%$\\

\bottomrule

\end{tabular}}
\caption{Zero-shot plan generation results using prompting-based approaches with Greedy decoding (average over all plans). Results with other metrics and standard deviation can be found in \S\ref{sec:appx-additional-results}.}

\label{tab:ablation-combined}
\end{table*}

\section{Data Collection}
\label{sec:data-collection}

In this section, we describe the data collection procedure for studying faithful plan generation. %

\noindent\textbf{Domain Creation.} Previous datasets have developed domains, flows, and APIs for tackling intents in TODs. We find, none of the existing datasets is usable as it is in our problem. For example, the STAR \cite{mosig2020star} and STARv2 \cite{zhao2023anytod} datasets contain APIs (without flows) to resolve user queries, however, dependencies among the APIs are sparse, e.g., STAR contains only $10$ dependencies among $25$ APIs. In real life, API dependencies are more dense, hence, these datasets are not very useful for our study where the goal is to study the ability of LLMs to preserve dependencies. The ABCD dataset \cite{chen2021action} contains flows for a few tasks, however, the related APIs needed to perform the steps in the flows are not given, hence, it is also not suitable for our study. Hence, inspired from these two datasets, we create in total, $4$ domains, $13$ flows (similar to ABCD) for resolving $13$ distinct intents in these domains, and $64$ APIs (similar to STAR) to execute different flow steps. We also annotate the gold plans to resolve the $13$ intents. The statistics of the dataset and examples are presented in Tab. \ref{tab:domain-dataset-summary}.

\noindent\textbf{Test Query Generation.} We aim to study plan generation in response to a user query in our created domains. To collect diverse user utterances related to the intents in Tab. \ref{tab:domain-and-flow-statistics}, we utilize the generative power of pretrained LLMs. We prompt an open-source LLM, \texttt{Falcon-40B-instruct} \cite{falcon40b}\footnote{We experimented with other open source LLMs and found \texttt{Falcon} to be the best in generating diverse utterances.}, to generate user queries related to the intents. We prompt the LLM to write paraphrases of a query with a specific intent. Prompt structure and generated examples are shown in \S\ref{sec:appx-test-utterance-generation}.
We manually go through the generated queries and discard any queries that are not related to the intent in the prompt. We find that $94\%$ of the generated queries are correct. In this manner, we generate $20$ user queries per intent resulting in $260$ queries in total. We use this dataset for studying flow-constrained plan generation given a user query.

\section{Constrained Decoding for Planning}
\label{sec:constrained-decoding}
In this section, first, we run an ablation on plan generation in various settings as explained in \S\ref{sec:zero-shot-planning-with-llms}, to understand the complexity of the task. Then we explain our proposed constrained decoding based algorithm for faithful plan generation.

\subsection{Ablation}
\label{sec:ablation}

To understand the complexity of zero-shot faithful planning in different settings, we run a zero-shot prompting-based ablation with a number of open-source LLMs pretrained with/without tools and of different parameter size. We use \texttt{Santacoder} \cite{allal2023santacoder}, \texttt{Falcon} \cite{refinedweb}, \texttt{Mpt} \cite{MosaicML2023Introducing}, \texttt{Mistral-v0.1} \cite{jiang2023mistral}%
, \texttt{Koala} \cite{koala_blogpost_2023}, \texttt{Vicuna} \cite{vicuna2023}, \texttt{Llama} \cite{touvron2023llama}, \texttt{ToolAlpaca} \cite{tang2023toolalpaca}, and \texttt{ToolLlaMA} \cite{qin2023toolllm}. The prompt structures for generating plans with and without thoughts are shown in \S\ref{sec:appx-prompt-structure-for-plan-generation}. 
The results for plan generation in the two settings are summarized in Tab. \ref{tab:ablation-combined}. Our main observation is that all LLMs fail heavily in zero-shot planning regardless of the setting, especially LLMs fail to preserve the API and flow step dependencies in plans. For example, the average percentage of inconsistent APIs in the two settings, ranges from $22.6\%$ to $49.6\%$. This finding is in line with previous studies on LLMs' capability in task planning and tool usage \cite{ruan2023tptu,valmeekam2022large}. We categorize our main observations below.

\begin{figure*}[ht!]
  \centering
  \includegraphics[width=1\textwidth]{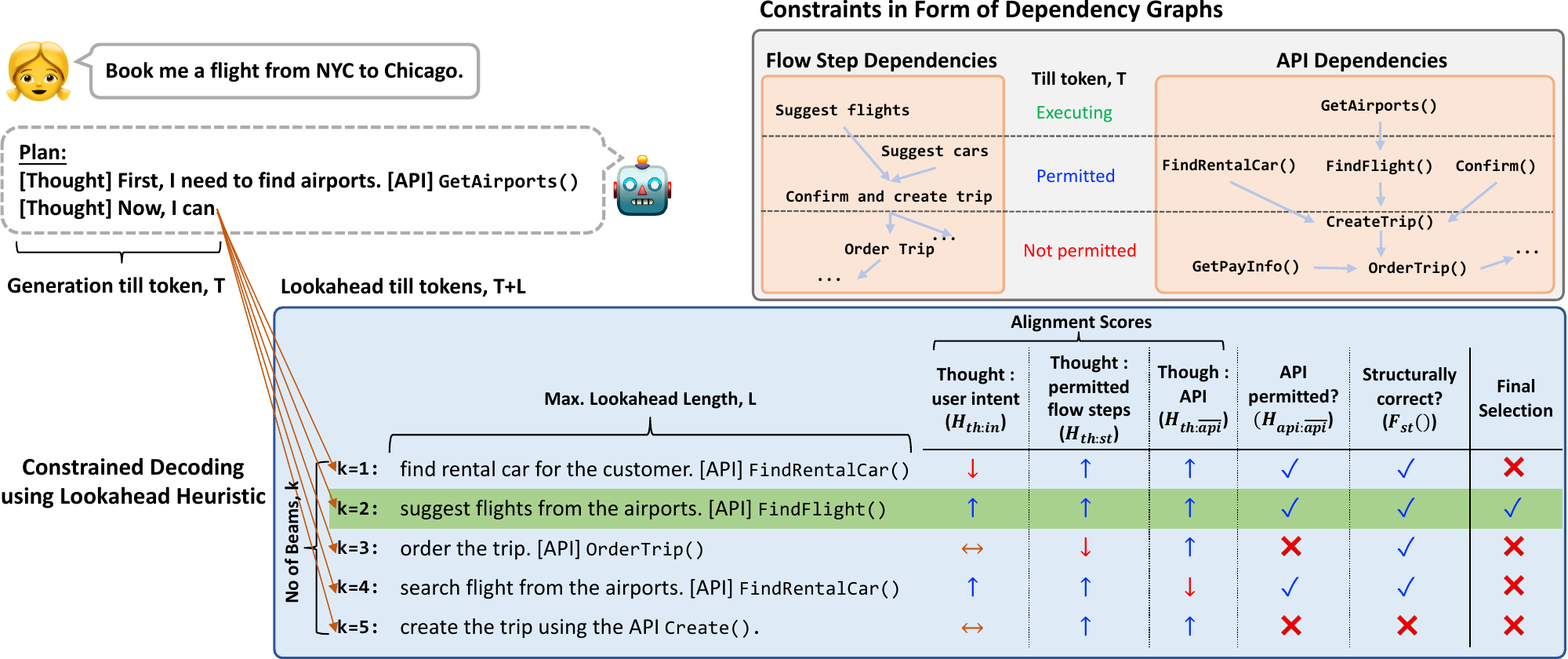}
  \caption{An example state of FLAP, our proposed lookahead heuristic-based constrained decoding algorithm for faithful planning, in the domain ``Trip Planning'' for a query related to ``book flight''. Here, \textcolor{blue}{$\uparrow$}, \textcolor{red}{$\downarrow$}, and \textcolor{orange}{$\leftrightarrow$} indicate high, low, and mediocre alignment scores, respectively. The selected path based on the heuristic scores is highlighted.} 
  \label{fig:constrained-decoding-example}
\end{figure*}

\noindent\textbf{Effect of thoughts.} Performance improves in flow faithfulness for almost all LLMs when prompted to generate thoughts and APIs compared to when prompted to generate only APIs in plan. This is consistent with the findings of \citet{yao2022react}. However, with thoughts, API repetition and hallucination increase and plan parsability (\S\ref{sec:appx-additional-results}) decreases a bit. On some test examples, we notice that the LLMs enter into a repetitive loop with thoughts (examples in \S\ref{repetition-error}), in line with well-known phenomenon in greedy and beam-search decoding \citep{vijayakumar2016diverse,shao2017generating}. 

\noindent\textbf{Effect of parameter size.} Performance in most of the metrics improve in larger LLMs and smaller LLMs are prone to hallucination. We observe that different models of the same parameter size perform differently. For example, the 7B version of \texttt{MPT} is better than the 7B version of \texttt{Falcon} in most of the metrics and \texttt{Mistral-7B} is better than other 7B models and it performs as per 30B-40B models. 

\noindent\textbf{Effect of pretraining.} We observe that LLMs pretrained on tools, e.g., \texttt{ToolAlpaca} fails heavily in API consistency and results in high API hallucination. With \texttt{ToolLlama}, we find that it fails to follow instructions and generates non-parsable responses (hence, not reported). These findings suggest that pretraining LLMs with tool data makes them heavily biased towards the pretraining data.

\noindent\textbf{All flows \textit{vs.} relevant flow in prompt.} When only the relevant flow to the user query is given in-context, the performance improves with almost all LLMs indicating that LLMs are more confused when multiple flows are given in-context. 

\subsection{Proposed Algorithm}
\label{sec:proposed-algorithm}

We now introduce \textbf{FLAP}, a \textbf{Fl}ow \textbf{A}dhering \textbf{P}lanning algorithm based on constrained decoding with lookahead heuristic (Fig. \ref{fig:constrained-decoding-example}). In FLAP, we first convert the flow steps and APIs to dependency graphs. At decoding time, we keep track of these dependency graphs to enforce the dependencies as constraints during next token generation. 

\subsubsection{Constructing Dependency Graphs}
For each domain, we construct dependency graphs for flow steps and APIs. The API dependency graph is constructed based on the input-output parameter dependencies among APIs. For example the API \texttt{FindFlight(inputs:airport; output- s:flights)} is dependent on the API \texttt{GetAirport (inputs:city; outputs:airport)} for the input parameter \texttt{`airport'}. Hence, in the API dependency graph \texttt{`GetAirport'} will be a parent of \texttt{`FindFlight'}. Similarly for flow steps, we construct a separate flow-step dependency graph. While generating a plan, we maintain a list of the steps and APIs that are executed in these dependency graphs. At each planning step, only non-executed/under-execution APIs/steps in the dependency graphs with parents in the executed list are considered permitted.

\subsubsection{Constrained Decoding of Plans with Lookahead Heuristic}
The traditional text generation problem can be formulated as, given an input sequence of tokens $x$ = $\{y_1, y_2, \cdots, y_{t-1}\}$, the next token generator predicts the next token $y_{t}$ by solving the equation:
\begin{equation}
y_{t} = \argmax_{y^{\prime}_{t}\in Y} P(y^{\prime}_{t}|x)\notag
\end{equation}
Here, $Y$ is the set of all possible generations and $P(y^{\prime}_{t}|x)$ is the likelihood of the token $y^{\prime}_{t}$ given the input sequence $x$. In constrained decoding, the likelihood function takes the form $P^{\prime}(y^{\prime}_{t}|x)$ = $P(y^{\prime}_{t}|x)$ + $H(y^{\prime}_{t}|x)$, where $H(y^{\prime}_{t}|x)$ is the score based on constraint satisfaction by generating $y^{\prime}_{t}$. %
However, often it is difficult to determine the constraint satisfaction score by only looking at the next token. The lookahead mechanism comes handy here %
where the constraint satisfaction score for a candidate next token, $y^{\prime}_{t}$ is estimated by looking at the estimated future generations if $y^{\prime}_{t}$ is selected. Hence, instead of $H(y^{\prime}_{t}|x)$, the heuristic function takes the form $H(y^{\prime}_{t}, \cdots, y^{\prime}_{t+L}|x)$, where $L$ is lookahead length.

For constrained plan generation, we prompt LLMs to generate thoughts and corresponding APIs. During generation, for each candidate next token, we lookahead till the end of one thought+API and score the candidate based on their constraint satisfaction. Hence, our heuristic function takes the form $H_{c}$ = $H([thought][api]|x)$. $[thought][api]$ may contain some already generated tokens; we also consider those when scoring. We score each $[thought][api]$ in the aspects below.

\paragraph{Generated thought to permitted flow step alignment ($H_{th:step}$).} In our approach, thoughts are natural language sentences describing the next step to be taken. Ideally, they should correspond to the flow steps. Hence, we score the generated thoughts based on their semantic similarity to the permitted flow steps in the flow dependency graph. A flow step is permitted, if all of its parent flow steps are already executed or the flow step is under execution (e.g., flow step consisting of multiple API calls). First, we estimate the intended flow step, $\hat{s}$ by a thought, $th$ by obtaining its semantically closest flow step using the formula below.
\begin{align}\small
    \hat{s} &= \argmax_{s\in S}sim(th, s)\notag
\end{align}
Here, $S$ is the set of all flow steps in the domain and $sim()$ is semantic similarity scorer. Then we calculate $H_{th:step}$ using the following equations.
\begin{align}\small
H_{th:step}&= 
\begin{cases}\small
    \text{\small $\alpha \times sim(th, \hat{s})$},& \text{\small if } \hat{s} \text{ \small in } S_{p}\notag\\
    \text{\small 0},              & \text{\small if } \hat{s} \text{ \small in } S_{n}\\
\end{cases}\\
    \alpha &= 
\begin{cases}\small
    \text{\small $\alpha_{a}$}, & \text{\small if }\hat{s}\text{ \small is in a flow} \notag\\
    \text{\small $\alpha_{b}$}, & \text{\small if }\hat{s}\text{ \small is deviating from flow}\\
    \text{\small $\alpha_{c}$}, & \text{\small if }\hat{s}\text{ \small is same as the last step}\\
\end{cases}
\end{align}
Here, $S_{p}$ and $S_{n}$ are permitted and not permitted flows steps at the current thought, respectively. We assign different weights to $H_{th:step}$ based on the intended flow step $\hat{s}$. If the intended flow step $\hat{s}$ belongs to the flow that the model has been following so far, we assign the weight $\alpha_{a}$. If the model deviates from the flow it has been following, we assign the weight $\alpha_{b}$. In single intent use-cases, the model is required to follow only one flow, hence, setting $\alpha_{a}$ > $\alpha_{b}$ prevents the model from deviating from the flow it has been following. If the model generates a flow step that is the same as the step it intended in its previous thought, we assign a weight $\alpha_{c}$. One step may involve executing multiple APIs. Hence, setting $\alpha_{c}$ > $\alpha_{a}$ encourages the model to complete the current step before moving to the next.  

\paragraph{Generated API to permitted APIs alignment ($H_{api:\overline{api}}$).} We score the generated API based on their alignment to the permitted APIs. Permitted APIs at a step are the non-executed APIs whose parents are already executed. We calculate $H_{api:\overline{api}}$ using the following equations. 
\begin{align}\small
    \hat{a} &= \argmax_{a\in A}sim(a, \overline{a})\notag\\
    H_{api:\overline{api}} &= \beta \times sim(\overline{a}, \hat{a})\notag\\
    \beta &= 
\begin{cases}\small
    \text{\small [0,1)}, & \text{\small if } \hat{a} \notin A_{p} \cup A_{c} \text{ \small or } \overline{a} \notin A\notag\\
    \text{\small 1},  & \text{\small if } \hat{a} \in A_{p}\\
    \text{\small 0},  & \text{\small if } \hat{a} \in A_{c}\\
\end{cases}
\end{align}
Here, $\overline{a}$ is the generated API, $A$ is the set of all APIs in a domain, $A_p$ and $A_c$ are permitted APIs and already called APIs, respectively, and $A_p \cap A_c = \emptyset$. When APIs from none of the sets $A_p$ and $A_c$ are generated, the API is either hallucinated or generated from the non-permitted list ($A_n$). In both of the cases, we assign a lower weight than 1 in the alignment score to make it a soft constraint while making it zero will enforce a hard constraint against such cases. We keep this weight open to tuning for different LLMs because of their varying degree of hallucinations and errors. 

\paragraph{Generated thought to user intent alignment ($H_{th:in}$).} The plan should correspond to the user intent in the query. Hence, we score the generated thought, $th$ based on its semantic similarity with the user query, $in$, using, $H_{th:in}$ = $sim(th, in)$.

\begin{table*}[ht!]
\centering

\begin{subtable}{2\columnwidth}\centering
{\resizebox{1\columnwidth}{!}{%
\begin{tabular}
{>{\centering\arraybackslash}m{1cm}|>{\centering\arraybackslash}m{0.2cm}>{\arraybackslash}m{8cm}>{\centering\arraybackslash}m{1.3cm}|>{\centering\arraybackslash}m{1.35cm}>{\centering\arraybackslash}m{1.35cm}|>{\centering\arraybackslash}m{2cm}|>{\centering\arraybackslash}m{1.7cm}>{\centering\arraybackslash}m{1.6cm}|>{\centering\arraybackslash}m{1.4cm}>{\centering\arraybackslash}m{1.4cm}|>{\centering\arraybackslash}m{1.6cm}>{\centering\arraybackslash}m{1.6cm}}
\toprule

& & & \textbf{With}	& \multicolumn{2}{c|}{\textbf{Avg count per plan}}	& \textbf{\% Plan} & \multicolumn{2}{c|}{\textbf{Avg \% of APIs in plan}} & \multicolumn{2}{c|}{\textbf{Avg \# of edit for}}	& \multicolumn{2}{c}{\textbf{Avg \% of inconsistent}}	\\

\textbf{LLMs} & & \textbf{Decoding Strategy} & \textbf{Thought}	& \textbf{Thoughts}	& \textbf{APIs}	& \textbf{Parsable ($\uparrow$)} & \textbf{Repeat ($\downarrow$)}	& \textbf{Hallu. ($\downarrow$)}	& \textbf{Steps ($\downarrow$)}	& \textbf{APIs ($\downarrow$)}	& \textbf{Steps ($\downarrow$)}	& \textbf{APIs ($\downarrow$)}\\
\toprule
         \multirow{8}{*}{\textbf{\rotatebox[origin=c]{90}{\texttt{mpt-7b-instruct}}}} & \multirow{4}{*}{\textbf{\rotatebox[origin=c]{90}{\textbf{Baselines}}}}  &  Greedy Search & No &	- &	$17.3$ &	$100\%$ &	$11.6\%$ &	$0.2\%$ &	$16.2$ &	$13.4$ &	$14.7\%$ &	$33.3\%$\\
                                                    & &  Greedy Search & Yes &	$16.2$ &	$16.1$ &	$97.7\%$ &	$35.1\%$ &	$4.0\%$ &	$13.9$ &	$12.4$ &	$11.9\%$ &	$29.6\%$\\
                                                    & & Beam Search &	Yes &	$9.9$ &	$9.9$ &	$70.8\%$ &	$5.5\%$ &	$3.0\%$ &	$7.6$ &	$7.7$ &	$13.9\%$ &	$41.0\%$\\
                                                    & & Nucleus Sampling &	Yes &	$12.3$ &	$12.4$ &	$76.9\%$ &	$25.3\%$ &	$4.0\%$ &	$9.8$ &	$9.5$ &	$13.8\%$ &	$31.7\%$\\
                                                    \cmidrule{2-13}
                                                    
                                                    & \multirow{4}{*}{\textbf{\rotatebox[origin=c]{90}{\textbf{Our model}}}}  &  FLAP.1 [soft api + align thought to (api, intent)]& Yes &	$10.3$ &	$10.4$ &	$100\%$ &	$1.3\%$ &	$0.0\%$ &	$5.5$ &	$5.5$ &	$6.3\%$ &	$1.3\%$\\
                                                    & &  FLAP.2 [soft api + align thought to (step, api)]& Yes &	$9.3$ &	$9.4$ &	$100\%$ &	$3.0\%$ &	$0.0\%$ &	$5.7$ &	$5.6$ &	$4.3\%$ &	$1.6\%$\\
                                                    & &  FLAP.3 [soft api + align thought to (step, intent)]& Yes &	$9.2$ &	$9.4$ &	$100\%$ &	$1.0\%$ &	$0.0\%$ &	$5.0$ &	$5.0$ &	$6.1\%$ &	$1.3\%$\\
                                                    & &  FLAP.4 [soft api + align thought to (step, api, intent)]& Yes &	$9.3$ &	$9.4$ &	$100\%$ &	$3.6\%$ &	$0.0\%$ &	$5.5$ &	$5.4$ &	$4.8\%$ &	$2.1\%$\\
         
         \midrule
         \multirow{8}{*}{\textbf{\rotatebox[origin=c]{90}{\texttt{mistral-7b-instruct}}}} & \multirow{4}{*}{\textbf{\rotatebox[origin=c]{90}{\textbf{Baselines}}}}  &  Greedy Search & No &	- &	$4.8$ &	$100\%$ &	$0.3\%$ &	$2.4\%$ &	$2.9$ &	$3.0$ &	$10.6\%$ &	$37.2\%$\\
                                                        & & Greedy Search & Yes &	$4.7$ &	$4.8$ &	$99.6\%$ &	$2.9\%$ &	$0.0\%$ &	$2.8$ &	$2.8$ &	$3.1\%$ &	$40.3\%$\\
                                                        & & Beam Search &	Yes &	$3.6$ &	$3.8$ &	$98.1\%$ &	$0.2\%$ &	$0.0\%$ &	$2.9$ &	$3.0$ &	$2.4\%$ &	$38.9\%$\\
                                                        & & Nucleus Sampling &	Yes &	$5.7$ &	$5.9$ &	$93.9\%$ &	$5.8\%$ &	$2.0\%$ &	$3.0$ &	$3.0$ &	$7.5\%$ &	$34.3\%$\\

                                                        \cmidrule{2-13}
                                                        
                                                        & \multirow{4}{*}{\textbf{\rotatebox[origin=c]{90}{\textbf{Our model}}}} & FLAP.1 [soft api + align thought to (api, intent)]& Yes &	$6.6$ &	$6.7$ &	$100\%$ &	$0.0\%$ &	$0.0\%$ &	$2.5$ &	$2.5$ &	$6.3\%$ &	$4.6\%$\\
                                                        & & FLAP.2 [soft api + align thought to (step, api)]& Yes &	$6.1$ &	$6.1$ &	$100\%$ &	$0.1\%$ &	$0.0\%$ &	$2.3$ &	$2.3$ &	$5.1\%$ &	$5.5\%$\\
                                                        & & FLAP.3 [soft api + align thought to (step, intent)]& Yes &	$6.3$ &	$6.4$ &	$100\%$ &	$0.4\%$ &	$0.0\%$ &	$2.0$ &	$2.0$ &	$6.5\%$ &	$3.6\%$\\
                                                        & & FLAP.4 [soft api + align thought to (step, api, intent)]& Yes &	$6.4$ &	$6.4$ &	$100\%$ &	$0.5\%$ &	$0.0\%$ &	$2.5$ &	$2.5$ &	$5.6\%$ &	$7.7\%$\\

\bottomrule

\end{tabular}}}

\centering
\caption{Setting: All flows in a domain are present in-context.}
\label{tab:cd-results-combined-all-flows}
\end{subtable}

\begin{subtable}{2\columnwidth}\centering
{\resizebox{1\columnwidth}{!}{%
\begin{tabular}
{>{\centering\arraybackslash}m{1cm}|>{\centering\arraybackslash}m{0.2cm}>{\arraybackslash}m{8cm}>{\centering\arraybackslash}m{1.3cm}|>{\centering\arraybackslash}m{1.35cm}>{\centering\arraybackslash}m{1.35cm}|>{\centering\arraybackslash}m{2cm}|>{\centering\arraybackslash}m{1.7cm}>{\centering\arraybackslash}m{1.6cm}|>{\centering\arraybackslash}m{1.4cm}>{\centering\arraybackslash}m{1.4cm}|>{\centering\arraybackslash}m{1.6cm}>{\centering\arraybackslash}m{1.6cm}}

    \multirow{8}{*}{\textbf{\rotatebox[origin=c]{90}{\texttt{mpt-7b-instruct}}}} & \multirow{4}{*}{\textbf{\rotatebox[origin=c]{90}{\textbf{Baselines}}}}  &  Greedy Search & No &	- &	$14.9$ &	$100\%$ &	$8.5\%$ &	$0.2\%$ &	$13.6$ &	$11.5$ &	$13.5\%$ &	$35.6\%$\\
                                                    & &  Greedy Search & Yes &	 $12.4$ &	$12.3$ &	$98.9\%$ &	$23.0\%$ &	$3.0\%$ &	$9.9$ &	$8.9$ &	$9.1\%$ &	$33.4\%$\\
                                                    & & Beam Search &	Yes &	$8.0$ &	$8.0$ &	$81.5\%$ &	$4.2\%$ &	$2.0\%$ &	$5.2$ &	$5.3$ &	$10.8\%$ &	$41.4\%$\\
                                                    & & Nucleus Sampling &	Yes &	$10.0$ &	$10.3$ &	$75.0\%$ &	$19.8\%$ &	$5.0\%$ &	$7.2$ &	$7.1$ &	$13.6\%$ &	$33.0\%$\\
    \cmidrule{2-13}
                                                    & \multirow{4}{*}{\textbf{\rotatebox[origin=c]{90}{\textbf{Our model}}}} &  FLAP.1 [soft api +  align thought to (api, intent)]& Yes &	$9.6$ &	$9.6$ &	$100\%$ &	$1.1\%$ &	$0.0\%$ &	$4.7$ &	$4.7$ &	$6.6\%$ &	$1.2\%$\\
                                                    & &  FLAP.2 [soft api +  align thought to (step, api)] & Yes &	$8.2$ &	$8.2$ &	$100\%$ &	$0.9\%$ &	$0.0\%$ &	$3.5$ &	$3.4$ &	$5.1\%$ &	$1.2\%$\\
                                                    & &  FLAP.3 [soft api +  align thought to (step, intent)]& Yes &	$8.8$ &	$9.0$ &	$100\%$ &	$1.2\%$ &	$0.0\%$ &	$4.2$ &	$4.2$ &	$6.3\%$ &	$2.4\%$\\
                                                    & &  FLAP.4 [soft api +  align thought to (step, api, intent)]& Yes &	$8.6$ &	$8.7$ &	$100\%$ &	$3.0\%$ &	$0.0\%$ &	$3.7$ &	$3.4$ &	$5.6\%$ &	$3.7\%$\\
    
    \midrule
    \multirow{8}{*}{\textbf{\rotatebox[origin=c]{90}{\texttt{mistral-7b-instruct}}}} & \multirow{4}{*}{\textbf{\rotatebox[origin=c]{90}{\textbf{Baselines}}}}   &  Greedy Search & No &	- &	$5.9$ &	$100\%$ &	$0.6\%$ &	$1.4\%$ &	$3.1$ &	$3.1$ &	$8.0\%$ &	$30.6\%$\\
                                                    & & Greedy Search  & Yes &	$5.6$ &	$5.7$ &	$100\%$ &	$3.8\%$ &	$1.0\%$ &	$2.7$ &	$2.6$ &	$2.8\%$ &	$34.4\%$\\
                                                    & & Beam Search &	Yes &	$4.2$ &	$4.3$ &	$100.0\%$ &	$0.0\%$ &	$0.0\%$ &	$2.4$ &	$2.5$ &	$1.5\%$ &	$36.7\%$\\
                                                    & & Nucleus Sampling &	Yes &	$5.6$ &	$5.9$ &	$98.1\%$ &	$3.7\%$ &	$1.0\%$ &	$2.9$ &	$3.0$ &	$6.78\%$ &	$30.2\%$\\
                                                    
    \cmidrule{2-13}
                                                    & \multirow{4}{*}{\textbf{\rotatebox[origin=c]{90}{\textbf{Our model}}}} & FLAP.1 [soft api +  align thought to (api, intent)]& Yes &	$7.0$ &	$7.0$ &	$100\%$ &	$0.1\%$ &	$1.0\%$ &	$2.7$ &	$2.8$ &	$6.0\%$ &	$4.1\%$\\
                                                    & & FLAP.2 [soft api +  align thought to (step, api)]& Yes &	$6.6$ &	$6.6$ &	$100\%$ &	$0.4\%$ &	$1.0\%$ &	$2.3$ &	$2.3$ &	$5.9\%$ &	$7.0\%$\\
                                                    & & FLAP.3 [soft api +  align thought to (step, intent)]& Yes &	$6.7$ &	$6.8$ &	$100\%$ &	$0.4\%$ &	$0.0\%$ &	$2.3$ &	$2.3$ &	$6.6\%$ &	$4.3\%$\\
                                                    & & FLAP.4 [soft api +  align thought to (step, api, intent)]& Yes &	$6.4$ &	$6.4$ &	$100\%$ &	$0.4\%$ &	$1.0\%$ &	$2.2$ &	$2.2$ &	$6.1\%$ &	$8.7\%$\\

\bottomrule

\end{tabular}}}
\caption{Setting: Only relevant flow to test query is present in-context.}
\label{tab:cd-results-combined-relevant-flows}
\end{subtable}

\caption{Comparison of our proposed approach, FLAP, with baselines in zero-shot faithful plan generation. Here, the numbers (FLAP.\#) indicate different ablation versions of FLAP. Structural constraint is applied in all versions of FLAP. Results with standard deviations can be found in Tables \ref{tab:cd-results-all-flows} and \ref{tab:cd-results-relevant-flows} in \S\ref{sec:appx-additional-results}. As reference for the ``Avg count per plan'' column, we note, there are $6.84$ APIs per gold plan.}

\label{tab:cd-results-combined}
\end{table*}

\paragraph{Generated thought to generated API alignment ($H_{th:\overline{api}}$).} We generate thought to resonate the API usage. Hence, the generated API, $\overline{a}$ should be relevant to the generated thought, $th$. We align the generated thought to the generated API using the score, $H_{th:\overline{api}}$ = $sim(th, \overline{a}_{d})$, where $\overline{a}_{d}$ is the given description of the generated API, $\overline{a}$. When $\overline{a}$ is a hallucination, we set $\overline{a}_{d}$ = $\overline{a}$.

\paragraph{Structural constraint ($F_{st}$).} For practical use-cases, it is convenient if the LLMs generate plans in a pre-defined format. It helps automatic parsability of the generated plans. Hence, we introduce a function, $F_{st}$ that accounts for the structural consistency of the generated plan. It operates on the combined heurictic score, $H_{c}^{\prime}$ = $a \times H_{th:step}$ + $b \times H_{api:\overline{api}}$ + $c \times H_{th:in}$ + $d \times H_{th:\overline{api}}$ ($a$, $b$, $c$, $d$ are scaling factors) as follows.
\begin{align}\small
    F_{st}(H_{c}^{\prime}) &= 
\begin{cases}\small
    \text{\small $H_{c}^{\prime}$},&\text{\small if thought+API follows format}\notag\\
    \text{\small 0},&\text{\small otherwise}
\end{cases}
\end{align}
We use $H_{c}$ = $F_{st}(H_{c}^{\prime})$ as the final heuristic score for a candidate token, $y_t^{\prime}$ and the final scoring function for next token selection is defined as follows.
\begin{align}\small
S(y_{t}^{\prime}|x) = (1-\lambda) \times P(y_{t}^{\prime}|x) + \lambda \times H_{c}^{y_{t}^{\prime}}\notag
\end{align}
The parameter $\lambda$ controls how much weight is given in the LLM logits, $P(y_{t}^{\prime}|x)$, and the heuristic score $H_{c}^{y_{t}^{\prime}}$. We use top-$k$ beams based on $P(y_{t}^{\prime})$ for the next token generation using constrained decoding. An example of our approach is shown in Fig. \ref{fig:constrained-decoding-example}.

\section{Experimental Evaluation}
\label{sec:experiments}

\subsection{Experimental Setting}
We evaluate FLAPs performance in faithful planning on the dataset described in \S\ref{sec:data-collection}, by comparing it with prompting-based baselines and different decoding methods. Note that, the approach with thoughts is similar to ReAct \cite{yao2022react}. Other approaches either depend on multiple modules, require finetuning, or tuned for a different task setting, hence, not directly applicable in our task. Based on our ablation (in Tab. \ref{tab:ablation-combined}), we apply FLAP on two LLMs in 7B parameter range, \texttt{Mistral-7b-instruct} and \texttt{Mpt-7b-instruct}. We leave application of FLAP on larger LLMs as future work because of their high resource requirement. We demonstrate the structure of plan, (\texttt{[thought]...[API]...}), through dummy start steps in the prompts (Fig. \ref{fig:prompt-for-plan-generation}, \S\ref{sec:appx-prompt-structure-for-plan-generation}) and enforce it through $F_{st}$ during decoding. We use a pretrained sentence transformer \cite{reimers2019sentence} to calculate semantic similarities ($sim()$). We report hyper-parameter tuning and other experimental details in \S\ref{sec:appx-experimental-details}.

\subsection{Experimental Results}
The experimental results are summarized in Table \ref{tab:cd-results-combined}. First, we observe that the Beam Search and Nucleus Sampling \cite{holtzman2019curious} methods do not improve the overall performance compared to the Greedy Search method. Rather both Beam Search and Nucleus Sampling result in fewer number of parsable plans compared to Greedy Search. Note that Beam Search and Nucleus Sampling explore a wider search space compared to Greedy Search, and Nucleus Sampling increases diversity; we observe that without constraints, it results in more deviation from the instructions and often decreases performance in different metrics. Hence, in the rest of this section, we compare our models performance with ``Greedy Search with thoughts''. We also observe that different baselines and our approach show similar comparative trends in the two settings (all flows in context vs. relevant flow in context). Hence, we focus our rest of the discussion on the more difficult setting where all flows are given in the prompt (Tab. \ref{tab:cd-results-combined-all-flows}).

In case of \texttt{Mpt}, different versions of our algorithm, FLAP, outperform the baselines by a great margin. In case of \texttt{Mistral}, the average percentage of inconsistent steps increases a bit (by $2$-$4$\%) with FLAP, however, performance improves in all other metrics, especially the inconsistent APIs decreases from $40.3\%$ to $3.6\%$. A similar trend is observed in the relevant flow in-context setting (Tab. \ref{tab:cd-results-combined-relevant-flows}). 

We observe that \texttt{Mistral} is good at following flows even without FLAP in the difficult all flows in-context setting, however, it fails to decompose when one flow step requires multiple API calls. Hence, without FLAP, the average API count per plan is lower ($4.8$) than the average number of APIs in gold plans ($6.84$). As a result, the avg. \% of inconsistent steps is low ($3.1$\%), however, \% of inconsistent API is very high ($40.3\%$). With FLAP, the decomposition capability improves and the model generates plans with number of APIs ($6.1$-$6.7$) closer to the gold reference value ($6.84$) with fewer number of edits and much less inconsistent APIs ($3.6$-$7.7\%$), however, with a trade off of slightly increased (by $2$-$4$\%) inconsistent steps. We present qualitative examples in Tab. \ref{tab:qualitative-decomposition-error-by-mistral}.

\begin{table}[t!]
\centering
\resizebox{1\columnwidth}{!}{%
\begin{tabular}
{>{\arraybackslash}m{15cm}}
{\textbf{Query: Can you book a flight from Boston to San Francisco?}}\\
\toprule
\textbf{Generated Plan by \texttt{Mistral-7b-instruct} \textcolor{red}{Without} FLAP}\\
$[$thought$]$ To suggest flights, I need to find flights from Boston to San Francisco. $[$API$]$ FindFlight()\\ 
$[$thought$]$ Once I have the flight details, I can confirm and create the trip. $[$API$]$ Confirm()\\ 
$[$thought$]$ After confirming the trip, I can order the trip. $[$API$]$ OrderTrip()\\ 
\midrule
\textbf{Generated Plan by \texttt{Mistral-7b-instruct} \textcolor{blue}{With} FLAP}\\
$[$thought$]$ First, I need to find the airport codes for Boston and San Francisco. $[$API$]$ GetAirports() \\
$[$thought$]$ Once I have the airport codes, I can suggest flights to the customer. $[$API$]$ FindFlight()\\
$[$thought$]$ After suggesting flights, I need to confirm the trip details with the user. $[$API$]$ Confirm()\\
$[$thought$]$ Once the trip details are confirmed, I can create the trip. $[$API$]$ CreateTrip()\\
$[$thought$]$ To order the trip, I need to get the payment information from the user. $[$API$]$ GetPayInfo()\\
$[$thought$]$ After getting the payment information, I can order the trip. $[$API$]$ OrderTrip()\\
\bottomrule
\end{tabular}}
\caption{Generated plans by \texttt{Mistral}. It often fails in step decomposition without FLAP.}
\label{tab:qualitative-decomposition-error-by-mistral}
\end{table}

We perform ablation by dropping different components of the heuristic scoring function in FLAP. We observe, different components of the heuristic score affect the corresponding aspects they are designed for, especially in the difficult setting where all flows are present in-context. For example, in case of \texttt{Mpt}, dropping $H_{th:step}$ (thought to flow step alignment) results in higher step dependency violation (FLAP.1). Another component $H_{th:in}$ (thought to intent alignment) is designed to keep the model focused on the flow related to the user intent and we observe an overall lower number of edits required when this component is added (FLAP.1, FLAP.3, FLAP.4), implying this component helps the model generate a plan related to the user intent. However, the effect of these components is not clear in case of \texttt{Mistral}. We conjecture, as \texttt{Mistral} is already good at following flows, the effect is lower. The effect of $H_{th:\overline{api}}$ that makes generated thoughts and APIs coherent, is more qualitative and we find that dropping these component (FLAP.3) results in inconsistent thought and API pairs even if the generated API is correct in the plan. Examples of such errors are presented in Tab. \ref{tab:effect-of-thought-api-alignmen}. Finally, enforcing structural consistency ($F_{st}$) during decoding makes the plans parsable $100\%$ of the time. 

\begin{figure}[t!]
    \centering
    \begin{subfigure}[t]{0.35\textwidth}
        \centering
        \resizebox{1\columnwidth}{!}{
        \includegraphics[trim={.1cm .1cm .1cm .1cm},clip]{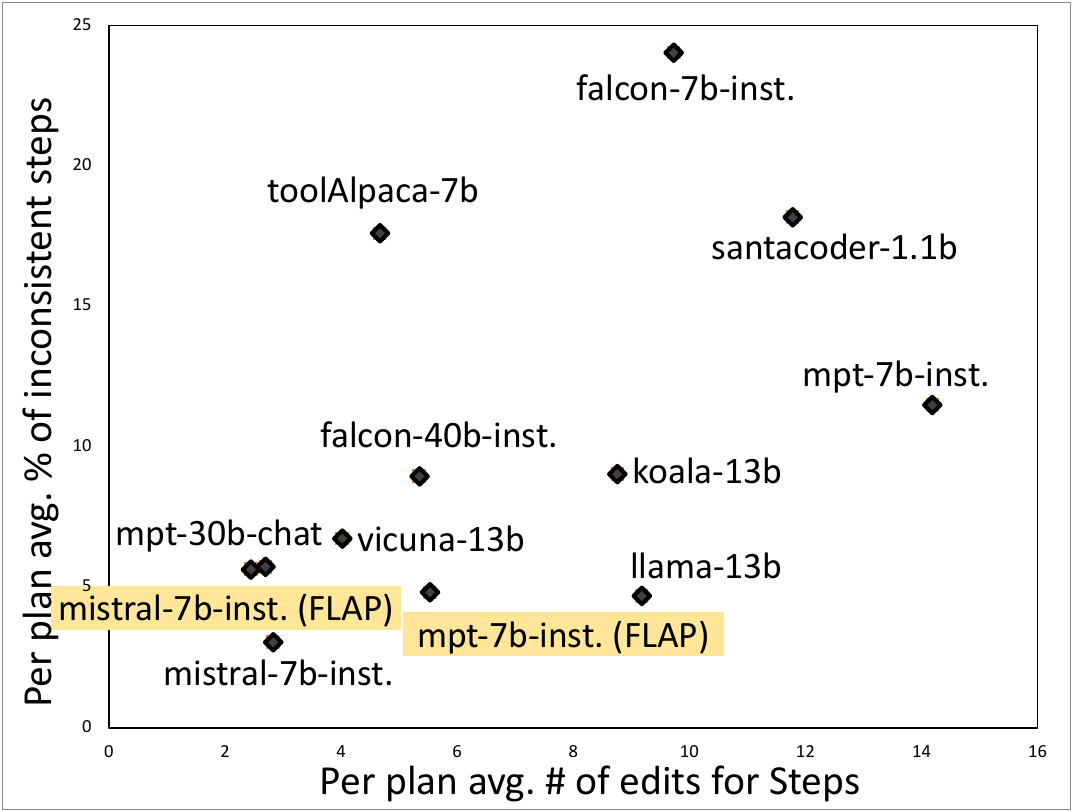}
        }
        \caption{Errors in generated flow steps.}
        \label{}
    \end{subfigure}\\
    \begin{subfigure}[t]{0.35\textwidth}
        \centering
        \resizebox{1\columnwidth}{!}{
        \includegraphics[trim={.1cm .1cm .1cm .1cm},clip]{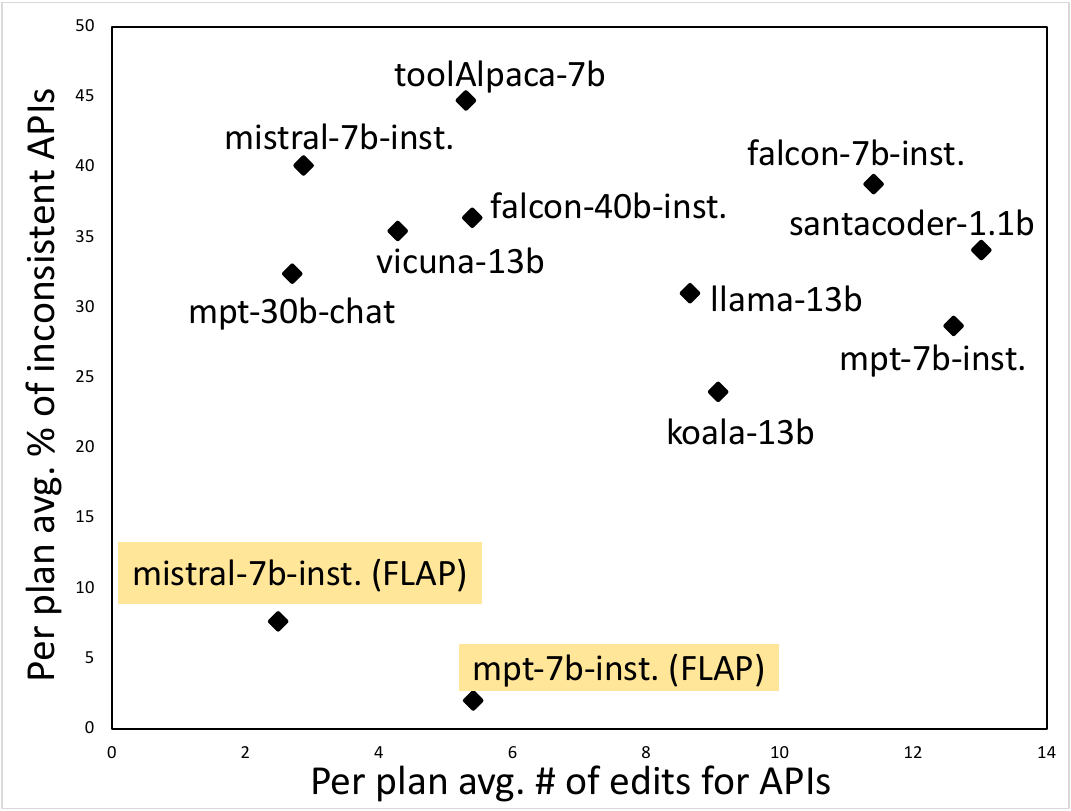}
        }
        \caption{Errors in generated APIs.}
        \label{}
    \end{subfigure}%
    
    \caption{Comparison among different LLMs in flow step (top) and API (bottom) errors during plan generation in the setting where all flows are given in context.}
    \label{fig:cd-comparison-with-larger-llms}
\end{figure}

In Fig. \ref{fig:cd-comparison-with-larger-llms}, we present the comparative performance of LLMs of different size and smaller 7B LLMs with FLAP. We observe that FLAP on top of \texttt{Mistral-7b} outperforms and \texttt{Mpt-7b} performs at par larger LLMs.

\begin{table*}[t]
\centering

\begin{subtable}{2\columnwidth}\centering
\resizebox{1\columnwidth}{!}{%
\begin{tabular}
{>{\arraybackslash}m{17cm}|>{\arraybackslash}m{16cm}}
\toprule
\multicolumn{2}{c}{\textbf{Query: I need to book a flight from NYC to Chicago.}}\\
\midrule
\textbf{Generated plan with FLAP \textcolor{blue}{with} thought to API alignment} & \textbf{Generated plan with FLAP \textcolor{red}{without} thought to API alignment}\\
\midrule
$[$thought$]$ I can get nearby airport codes to suggest flights to the customer. $[$API$]$ GetAirports() & \colorbox{pink}{$[$thought$]$ The customer wants to book a flight from NYC to Chicago. $[$API$]$ GetAirports()}\\
$[$thought$]$ I can suggest flights to the customer. $[$API$]$ FindFlight() & $[$thought$]$ I can suggest flights to the customer. $[$API$]$ FindFlight()\\
$[$thought$]$ I can confirm flights to the customer and create a trip. $[$API$]$ Confirm() & \colorbox{pink}{$[$thought$]$ I have suggested flights to the customer. $[$API$]$ GetPayInfo()}\\
$[$thought$]$ I confirm flights to the customer and create a trip. $[$API$]$ CreateTrip() & $[$thought$]$ I can confirm and create the trip. $[$API$]$ Confirm()\\
$[$thought$]$ Before ordering the trip, I can get payment information from the customer. $[$API$]$ GetPayInfo() & $[$thought$]$ Confirms and create the trip. $[$API$]$ CreateTrip()\\
$[$thought$]$ I confirm flights to the customer and order a trip. $[$API$]$ OrderTrip() & \colorbox{pink}{$[$thought$]$ The trip is created. $[$API$]$ OrderTrip()}\\
\bottomrule
\end{tabular}}
\caption{Generated plan with \texttt{Mpt-7b-instruct}.}
\end{subtable}

\begin{subtable}{2\columnwidth}\centering
\resizebox{1\columnwidth}{!}{%
\begin{tabular}
{>{\arraybackslash}m{17cm}|>{\arraybackslash}m{16cm}}
$[$thought$]$ In order to suggest flights to the customer, I need to get the airport codes. $[$API$]$ GetAirports() & $[$thought$]$ In order to suggest flights to the customer, I need to get the airport codes. $[$API$]$ GetAirports()\\
$[$thought$]$ Once I have the airport codes, I can suggest flights to the customer. $[$API$]$ FindFlight() & $[$thought$]$ Once I have the airport codes, I can suggest flights to the customer. $[$API$]$ FindFlight()\\
$[$thought$]$ After suggesting the flights, I need to confirm and create the trip. $[$API$]$ Confirm() & $[$thought$]$ After suggesting the flights, I need to confirm and create the trip. $[$API$]$ Confirm()\\
$[$thought$]$ To order the trip, I need to create a trip with the suggested flight. $[$API$]$ CreateTrip() & \colorbox{pink}{$[$thought$]$, $[$API$]$ CreateTrip()}\\
$[$thought$]$ Before ordering the trip, I need to get the payment information from the customer. $[$API$]$ GetPayInfo() & \colorbox{pink}{$[$thought$]$ Once the trip is created, I can order the trip. $[$API$]$ GetPayInfo()}\\
$[$thought$]$ Once I have the payment information, I can order the trip. $[$API$]$ OrderTrip() & $[$thought$]$, order the trip. $[$API$]$ OrderTrip()\\
\bottomrule
\end{tabular}}
\caption{Generated plan with \texttt{Mistral-7b-instruct}.}
\end{subtable}

\caption{Qualitative error analysis when the generated thought to generated API alignment component ($H_{th:\overline{api}}$) in the heuristic scoring function of FLAP is dropped. The models results in incoherent thoughts and APIs at steps if $H_{th:\overline{api}}$ is dropped; such steps are highlighted in red.}
\label{tab:effect-of-thought-api-alignmen}
\end{table*}  

\section{Related Work}
\label{sec:related-works}

While API usage and planning have been studied as separate concepts in literature, we argue that these two are interdependent. %
Recently, different approaches have been proposed to teach LLMs to use the correct API, however, the faithfulness of API usage to their dependencies is understudied \cite{ruan2023tptu}. For correct API usage, one line of research considers pretraining of LLMs with large API data that are mostly augmented using self-instruction \citep{schick2023toolformer,patil2023gorilla,qin2023toolllm,yang2023gpt4tools,gao2023confucius,tang2023toolalpaca,li2023api}. In some cases, authors considered modular approaches and relied on feedback from humans, other LLMs, or API execution outputs to tune LLMs \citep{song2023restgpt,qin2023tool,qiao2023making,lu2023gear,prasad2023adapt,chen2023chatcot}. Another line of work focused on improving the reasoning mechanism \cite{qian2023creator} for API usage via prompting to generate thoughts \cite{yao2022react}, consider API documentation \cite{hsieh2023tool}, etc. Unfortunately, pretraining-based methods are difficult and prohibitively costly when adaptation to new APIs is required and prompting-based approaches cannot reliably improve reasoning and planning which are key to faithful API usage~(\S\ref{sec:ablation}).

In literature, planning is defined as decomposition of the end task and identifying intermediate executable actions or reasoning steps to achieve the goal where the decomposition is open-ended and does not require obeying instructions \citep{jansen2020visually,huang2022language,ahn2022can,wu2022understanding,lu2022neuro,yuprompt,lu2023chameleon,brahman2023plasma}. However, in many applications, such as customer care service or physical robots, planning needs to strictly adhere to predefined instructions \cite{reijers2003design,sengupta2019monitor}. To achieve this, we draw inspiration from decoding approaches \citep{holtzman2019curious,meister-etal-2020-best} that enable enforcement of lexical and parametric constraints during generation \citep{hokamp2017lexically,lu2021neurologic,lu2022neurologic,krause2021gedi,yang2021fudge,huang2024deal}. We extend the idea to enforce constraints related to predefined instructions.

\section{Conclusion}
\label{sec:conclusion}

In this paper, we study the novel problem of faithful planning for performing tasks in TODs, that adheres to predefined flows and preserves API dependencies. To solve the task, we propose \textbf{FLAP}, a constrained decoding algorithm based on lookahead heuristic. FLAP outperforms prompting-based baselines with different decoding methods and applying FLAP on two smaller LLMs we achieve comparable performance to larger ones. Our approach can be leveraged in accurate planning for various downstream applications such as next step prediction in conversations, user simulation, flow-following conversation data augmentation for training TOD models, and so on.

\section*{Limitations}
\label{sec:limitations}
We identify the following limitations of our study.

\paragraph{Simplifying Assumptions:} In our study, we make the simplifying assumption that the user query can be resolved using the given APIs and workflows in context. However, that may not be the case in real life and users may have out-of-domain (OOD) queries. To tackle OOD intents, we can easily adapt our model by adding one additional workflow such as ``cannot help'', and instruct it to be selected if no other workflow is similar to the user query. Our constructed dataset also do not contain cases where multiple plans are valid based on the given workflows and APIs, however, our algorithm can be applicable in such cases without any modification.

\paragraph{Static Planning:} In this paper, we study static planning. However, in real life, the agents may need to modify their plans in between a conversation because conversations are by nature dynamic. Some possible reasons for plan adaptation may be, based on API execution feedback, user changes their mind in between, and so on. Adapting such static plans based on the situation in dynamic conversations can be an interesting future work. We also focus only on API selection by following pre-defined flows and API dependencies. The correctness of parameter filling for the APIs is left as a future work as it is applicable mostly in the dynamic planning setting. 

\paragraph{Runtime:} We observe a higher runtime for plan generation using constrained decoding (\S\ref{sec:appx-runtime-analysis}). This is partly because of the inefficient deployment of the LLMs in the existing implementations (we use Huggingface implementations) and partly because of our resource constraint (discussed in details in \S\ref{sec:appx-runtime-analysis}). We intend to work on more efficient implementation of our algorithm by potentially parallelizing computations and implementing algorithmic optimizations (e.g., lookahead after every $n$ tokens generation).

\paragraph{Usage of Open Source LLMs:} We did not use any closed source LLMs for our research because of various reasons. Firstly, our algorithm requires the logits from the LLMs which may be difficult to obtain using closed source LLMs. Secondly, they might be changing inside the box and it will be difficult to replicate the results, hence, harming the benchmarking procedure. Thirdly, closed source LLMs are costly as they are accessible through paid APIs only. Due to resource constraint, we experimented with LLMs of at most 40B parameter size. However, as described in the paper, our approach can be applied on top of any autoregressive LLM.

\section*{Ethics Statement}
\label{sec:ethics}
In this paper, we present all implementation and dataset details to replicate the study (partially in the Appendix). All the relevant datasets, publicly available LLMs, and other models used in this paper are publicly available for scientific research and are cited adequately. All the results are reported with standard deviations and necessary error analyses are done (some parts in the Appendix), to provide the readers an idea about the potential error patterns and risks related to using our proposed models. %

\section*{Acknowledgements}
\label{sec:acknowledgements}
We gratefully acknowledge James Gung, Yi-An Lai, Nikolaos Pappas, and the members of the AWS AI Labs for providing valuable feedback at different stages of this work. We extend our special thanks to Justin Sun for his help in setting up different LLMs and Don Bennett for providing valuable feedback on writing. We would also like to thank the anonymous reviewers for their insightful comments.

\bibliography{anthology,custom}
\bibliographystyle{acl_natbib}

\appendix

\section{Dataset Creation}
\label{sec:appx-dataset-creation}

\subsection{Domain Creation}
\label{sec:appx-domain-creation}
The flows and corresponding APIs for each domain are shown in Tab. \ref{tab:appx-domain-and-flow-example}. The API definitions in all domains, their input/output parameters, and descriptions are shown in Tab. \ref{tab:api-definitions}.

\begin{table*}[h]
\centering
\resizebox{1.5\columnwidth}{!}{%
\begin{tabular}
{>{\arraybackslash}m{8.5cm}|>{\arraybackslash}m{13cm}}
\toprule
\textbf{Flow steps} & \textbf{Required API sequence}\\
\toprule
\multicolumn{2}{c}{\textbf{Domain: Trip Booking, Intent: Book Car}}\\
\midrule
Start processing the requests from the customer & \texttt{InitSystem(), Start()}\\
Suggest cars to the customer                    & \texttt{FindRentalCar()}\\
Confirm and create the trip                     & \texttt{Confirm(), CreateTrip()}\\
Extract and add promotional offers              & \texttt{GetCarInsuranceDiscount(), UpdateTrip()}\\
Order the trip                                  & \texttt{GetPaymentInformation(), OrderTrip()}\\
Finish processing request                       & \texttt{Finish()}\\
\midrule

\multicolumn{2}{c}{\textbf{Domain: Trip Booking, Intent: Book Flight}}\\
\midrule
Start processing the requests from the customer & \texttt{InitSystem(), Start()}\\
Suggest flights to the customer                 & \texttt{GetAirports(), FindFlight()}\\
Confirm and create the trip                     & \texttt{Confirm(), CreateTrip()}\\
Order the trip                                  & \texttt{GetPaymentInformation(), OrderTrip()}\\
Finish processing request                       & \texttt{Finish()}\\
\midrule

\multicolumn{2}{c}{\textbf{Domain: Trip Booking, Intent: Book Hotel}}\\
\midrule
Start processing the requests from the customer & \texttt{InitSystem(), Start()}\\
Suggest hotels to the customer                  & \texttt{FindHotel()}\\
Confirm and create the trip                     & \texttt{Confirm(), CreateTrip()}\\
Order the trip                                  & \texttt{GetPaymentInformation(), OrderTrip()}\\
Finish processing request                       & \texttt{Finish()}\\
\midrule

\multicolumn{2}{c}{\textbf{Domain: Insurance, Intent: Buy Insurance}}\\
\midrule
Start processing the requests from the customer & \texttt{InitSystem(), Start()}\\
Provide quote for the item to be insured    & \texttt{GetItem(), GetQuote()}\\
Get customer details                        & \texttt{GetDemographicDetails()}\\
Order the insurance                         & \texttt{OrderInsurance()}\\
Finish processing request                       & \texttt{Finish()}\\
\midrule

\multicolumn{2}{c}{\textbf{Domain: Insurance, Intent: Cancel Insurance}}\\
\midrule
Start processing the requests from the customer & \texttt{InitSystem(), Start()}\\
Validate insurance                          & \texttt{GetInsuranceID(), ValidateInsurance()}\\
Record the refund method                    & \texttt{GetRefundMethod()}\\
Confirm and cancel the Insurance            & \texttt{Confirm(), CancelInsurance()}\\
Finish processing request                       & \texttt{Finish()}\\
\midrule

\multicolumn{2}{c}{\textbf{Domain: Insurance, Intent: Add Member}}\\
\midrule
Start processing the requests from the customer & \texttt{InitSystem(), Start()}\\
Validate insurance                                  & \texttt{GetInsuranceID(), ValidateInsurance()}\\
Update the insurance with new member information    & \texttt{GetAdditionalMemberInformation(), UpdateInsurance()}\\
Finish processing request                       & \texttt{Finish()}\\
\midrule

\multicolumn{2}{c}{\textbf{Domain: Banking, Intent: Open Account}}\\
\midrule
Start processing the requests from the customer & \texttt{InitSystem(), Start()}\\
Check eligibility for opening an account        & \texttt{GetCustomerIncome(), GetDateOfBirth(), CheckEligibility()}\\
Finalize the account type                       & \texttt{GetAccountTypes()}\\
Open the account                                & \texttt{Confirm(), OpenAccount()}\\
Finish processing request                       & \texttt{Finish()}\\
\midrule

\multicolumn{2}{c}{\textbf{Domain: Banking, Intent: Report Problem}}\\
\midrule
Start processing the requests from the customer & \texttt{InitSystem(), Start()}\\
Record the problem and notify the internal team & \texttt{RecordIssue(), NotifyComplaintDepartment()}\\
Finish processing request                       & \texttt{Finish()}\\
\midrule

\multicolumn{2}{c}{\textbf{Domain: Banking, Intent: Cancel Transaction}}\\
\midrule
Start processing the requests from the customer & \texttt{InitSystem(), Start()}\\
Validate transaction details    & \texttt{GetTransactionInformation(), ValidateTransaction()}\\
Cancel the transaction          & \texttt{Confirm(), CancelTransaction()}\\
Finish processing request                       & \texttt{Finish()}\\
\midrule

\multicolumn{2}{c}{\textbf{Domain: Restaurant \& Ride Book, Intent: Book Restaurant}}\\
\midrule
Start processing the requests from the customer & \texttt{InitSystem(), Start()}\\
Suggest restaurants to the customer     & \texttt{GetTime(), GetCity(), GetBudget(), FindRestaurants()}\\
Finalize the restaurant booking         & \texttt{GetCustomerName(), GetSelectedRestaurant(), BookRestaurant()}\\
Finish processing request                       & \texttt{Finish()}\\
\midrule

\multicolumn{2}{c}{\textbf{Domain: Restaurant \& Ride Book, Intent: Book Ride}}\\
\midrule
Start processing the requests from the customer & \texttt{InitSystem(), Start()}\\
Suggest ride options to the customer    & \texttt{GetTime(), GetSource(), GetDestination(), FindRideShareOptions()}\\
Finalize the ride service booking       & \texttt{GetPaymentMethod(), GetSelectedRideShareOption(), BookRide()}\\
Finish processing request                       & \texttt{Finish()}\\
\midrule

\multicolumn{2}{c}{\textbf{Domain: Restaurant \& Ride Book, Intent: Cancel Restaurant Booking}}\\
\midrule
Start processing the requests from the customer & \texttt{InitSystem(), Start()}\\
Validate the restaurant booking     & \texttt{GetRestaurantBookingID(), Authenticate()}\\
Cancel the restaurant booking       & \texttt{CancelRestaurantBooking()}\\
Finish processing request                       & \texttt{Finish()}\\
\midrule

\multicolumn{2}{c}{\textbf{Domain: Restaurant \& Ride Book, Intent: Cancel Ride Booking}}\\
\midrule
Start processing the requests from the customer & \texttt{InitSystem(), Start()}\\
Validate the ride booking       & \texttt{GetRideBookingID(), Authenticate()}\\
Cancel the ride booking         & \texttt{GetRefundMethod(), CancelRide()}\\
Finish processing request                       & \texttt{Finish()}\\
\bottomrule

\end{tabular}}
\caption{Flows and required APIs in all domains. We study faithful plan generation in a zero-shot setting. Hence, we additionally add the dummy first (\texttt{InitSystem(), Start()}) and last (\texttt{Finish()}) steps in each flow to guide the model by indicating the start and end of processing a request in the zero-shot prompt.}
\label{tab:appx-domain-and-flow-example}
\end{table*}

\begin{table*}[h!]
\centering
\resizebox{1.8\columnwidth}{!}{%
\begin{tabular}
{>{\arraybackslash}m{5.5cm}|>{\arraybackslash}m{5.2cm}|>{\arraybackslash}m{6cm}|>{\arraybackslash}m{10cm}}
\toprule
\textbf{APIs} & \textbf{Input Parameters} & \textbf{Output Parameters} & \textbf{Description}\\
\toprule

\multicolumn{4}{l}{\textbf{Domain: Trip Booking}}\\
\midrule
\texttt{Confirm()} & \texttt{None} & \texttt{confirmation\_status} & confirm trip details with the customer before creating a trip\\
\texttt{FindRentalCar()} & \texttt{None} & \texttt{car\_id} & retrieves cars to/from the travelling city\\
\texttt{FindHotel()} & \texttt{None} & \texttt{hotel\_id} & finds and retrieves hotels in a city\\
\texttt{GetPayInfo()} & \texttt{None} & \texttt{pay\_info} & gets payment information from the customer\\
\texttt{Start()} & \texttt{init\_status} & \texttt{True/False} & starts processing customer requests\\
\texttt{InitSystem()} & \texttt{None} & \texttt{init\_status} & initializes the system to start processing customer requests\\
\texttt{OrderTrip()} & \texttt{trip\_id,pay\_info, confirmation\_status} & \texttt{order\_status} & places the bookings\\
\texttt{UpdateTrip()} & \texttt{offer\_id, trip\_id} & \texttt{True/False} & updates a trip with special offers\\
\texttt{FindFlight()} & \texttt{airport\_code} & \texttt{flight\_id} & retrieves flights to/from the airport\\
\texttt{GetCarInsuranceDiscount()} & \texttt{car\_id} & \texttt{offer\_id} & retrieves insurances related discounts related to the car options\\
\texttt{CreateTrip()} & \texttt{flight\_id/hotel\_id/car\_id, confirmation\_status} & \texttt{trip\_id} & creates a new trip before ordering\\
\texttt{Finish()} & \texttt{order\_status} & \texttt{True/False} & finishes processing customer requests\\
\texttt{GetAirports()} & \texttt{None} & \texttt{airport\_code} & returns nearby airport codes based on the travelling city\\
\midrule

\multicolumn{4}{l}{\textbf{Domain: Insurance}}\\
\midrule
\texttt{CancelInsurance()} & \texttt{insurance\_id, confirmation\_status, refund\_method} & \texttt{cancellation\_status} & cancels the insurance\\
\texttt{GetAdditionalMemberInfo()} & \texttt{None} & \texttt{additional\_member\_info} & asks additional member info from the customer and returns it\\
\texttt{GetInsuranceID()} & \texttt{None} & \texttt{insurance\_id} & asks the insurance id from the customer and returns it\\
\texttt{OrderInsurance()} & \texttt{pay\_info} & \texttt{order\_status} & places order for purchasing the insurance\\
\texttt{GetPaymentInformation()} & \texttt{None} & \texttt{pay\_info} & gets payment information from the customer\\
\texttt{Finish()} & \texttt{order\_status/ cancellation\_status/ update\_status} & \texttt{True/False} & finishes processing customer requests\\
\texttt{GetRefundMethod()} & \texttt{None} & \texttt{refund\_method} & asks the preferred refund method from the customer and returns it\\
\texttt{Confirm()} & \texttt{insurance\_id} & \texttt{confirmation\_status} & confirms insurance details with the customer\\
\texttt{GetItem()} & \texttt{None} & \texttt{item\_id} & gets the item to be insured from the customer\\
\texttt{InitSystem()} & \texttt{None} & \texttt{init\_status} & initializes the system to start processing customer requests\\
\texttt{GetDemographicDetails()} & \texttt{None} & \texttt{demographics} & asks demographic information of the customer and returns it\\
\texttt{Start()} & \texttt{init\_status} & \texttt{True/False} & starts processing customer requests\\
\texttt{GetQuote()} & \texttt{item\_id} & \texttt{quote\_id} & provides quote to the customer for a given item\\
\texttt{ValidateInsurance()} & \texttt{insurance\_id} & \texttt{insurance\_validated} & validates if the insurance exists\\
\texttt{UpdateInsurance()} & \texttt{additional\_member\_info, insurance\_id} & \texttt{update\_status} & updates an existing insurance by adding a new member\\
\midrule

\multicolumn{4}{l}{\textbf{Domain: Banking}}\\
\midrule
\texttt{Start()} & \texttt{init\_status} & \texttt{True/False} & starts processing customer requests\\
\texttt{GetTransactionInfo()} & \texttt{None} & \texttt{transaction\_id} & asks transaction info to the customer and returns the transaction id\\
\texttt{CheckEligibility()} & \texttt{date\_of\_birth, annual\_income} & \texttt{eligibility\_status} & checks if customer is eligible to open account given age, income\\
\texttt{GetAccountTypes()} & \texttt{date\_of\_birth,annual\_income, eligibility\_status} & \texttt{account\_type} & get the account type based on customer eligibility, age and income\\
\texttt{CancelTransaction()} & \texttt{transaction\_id, confirmation\_status} & \texttt{cancellation\_status} & cancels a transaction\\
\texttt{ValidateTransaction()} & \texttt{transaction\_id} & \texttt{None} & validates if the transaction was actually executed\\
\texttt{Finish()} & \texttt{open\_status/ cancellation\_status/ notification\_status} & \texttt{True/False} & finishes processing customer requests\\
\texttt{GetCustomerIncome()} & \texttt{None} & \texttt{annual\_income} & ask customer annual income and returns it\\
\texttt{RecordIssue()} & \texttt{None} & \texttt{issue\_id} & records the issue in internal database for review by team\\
\texttt{OpenAccount()} & \texttt{confirmation\_status, account\_type} & \texttt{open\_status} & opens a new account\\
\texttt{InitSystem()} & \texttt{None} & \texttt{init\_status} & initializes the system to start processing customer requests\\
\texttt{Confirm()} & \texttt{None} & \texttt{confirmation\_status} & confirms with the customer before opening an account or cancelling a transaction\\
\texttt{GetDateOfBirth()} & \texttt{None} & \texttt{date\_of\_birth} & asks the date of birth to the customer and returns it\\
\texttt{NotifyComplaintDepartment()} & \texttt{issue\_id} & \texttt{notification\_status} & notifies the complaint department about an problem reported by the customer\\
\midrule

\multicolumn{4}{l}{\textbf{Domain: Restaurant \& Ride Booking}}\\
\midrule
\texttt{GetBudget()} & \texttt{None} & \texttt{budget} & asks customer the budget and return it\\
\texttt{FindRideShareOptions()} & \texttt{source, destination, time} & \texttt{ride\_options} & retrieves ride-share options\\
\texttt{GetPaymentMethod()} & \texttt{None} & \texttt{pay\_method} & asks customer the payment method and return it\\
\texttt{GetDestination()} & \texttt{None} & \texttt{destination} & asks destination to the customer and return it\\
\texttt{GetCustomerName()} & \texttt{None} & \texttt{name} & asks customer their name and return it\\
\texttt{BookRide()} & \texttt{pay\_method, rideshare\_id} & \texttt{ride\_booking\_status} & books the selected rideshare service\\
\texttt{InitSystem()} & \texttt{None} & \texttt{init\_status} & initializes the system to start processing customer requests\\
\texttt{GetSelectedRestaurant()} & \texttt{restaurant\_list} & \texttt{restaurant\_id} & presents customer with the list of restaurants, ask their choice and return the selected restaurant id\\
\texttt{Authenticate()} & \texttt{ride\_id/restaurant\_book\_id} & \texttt{True/False} & authenticates the customer and their booking ids\\
\texttt{Finish()} & \texttt{None} & \texttt{ride\_booking\_status/ restaurant\_booking\_status/ restaurant\_cancellation\_status/ ride\_cancellation\_status} & finishes processing customer requests\\
\texttt{GetRestaurantBookingID()} & \texttt{None} & \texttt{restaurant\_book\_id} & asks the customer the restaurant booking id and return it\\
\texttt{GetCity()} & \texttt{None} & \texttt{city} & asks customer the city and return it\\
\texttt{GetTime()} & \texttt{None} & \texttt{time} & asks time to the customer and return it\\
\texttt{GetSelectedRideShareOption()} & \texttt{ride\_options} & \texttt{rideshare\_id} & presents customer with the list of rides, asks their choice and return the selected rideshare id\\
\texttt{GetSource()} & \texttt{None} & \texttt{source} & asks source to the customer and return it\\
\texttt{BookRestaurant()} & \texttt{restaurant\_id, time, name} & \texttt{restaurant\_booking\_status} & books the selected restaurant\\
\texttt{CancelRide()} & \texttt{refund\_method, ride\_id} & \texttt{ride\_cancellation\_status} & cancels the ride\\
\texttt{FindRestaurants()} & \texttt{time, city, budget} & \texttt{restaurant\_list} & retrieves list of restaurants\\
\texttt{CancelRestaurantBooking()} & \texttt{restaurant\_book\_id} & \texttt{restaurant\_cancellation\_status} & cancels the restaurant booking\\
\texttt{Start()} & \texttt{init\_status} & \texttt{True/False} & starts processing customer requests\\
\texttt{GetRideBookingID()} & \texttt{None} & \texttt{ride\_id} & asks the customer the ride booking id and return it\\
\texttt{GetRefundMethod()} & \texttt{None} & \texttt{refund\_method} & asks customer the refund method and return it\\

\bottomrule

\end{tabular}}
\caption{API definition, their input/output parameters, and descriptions in four constructed domains. Some API names are clipped for better visibility (e.g., \texttt{GetAdditionalMemberInformation()} is clipped to \texttt{GetAdditionalMemberInfo()}).}
\label{tab:api-definitions}
\end{table*}

\subsection{Test Utterance Generation}
\label{sec:appx-test-utterance-generation}
\begin{figure*}[h]
  \centering
  \includegraphics[width=0.8\textwidth]{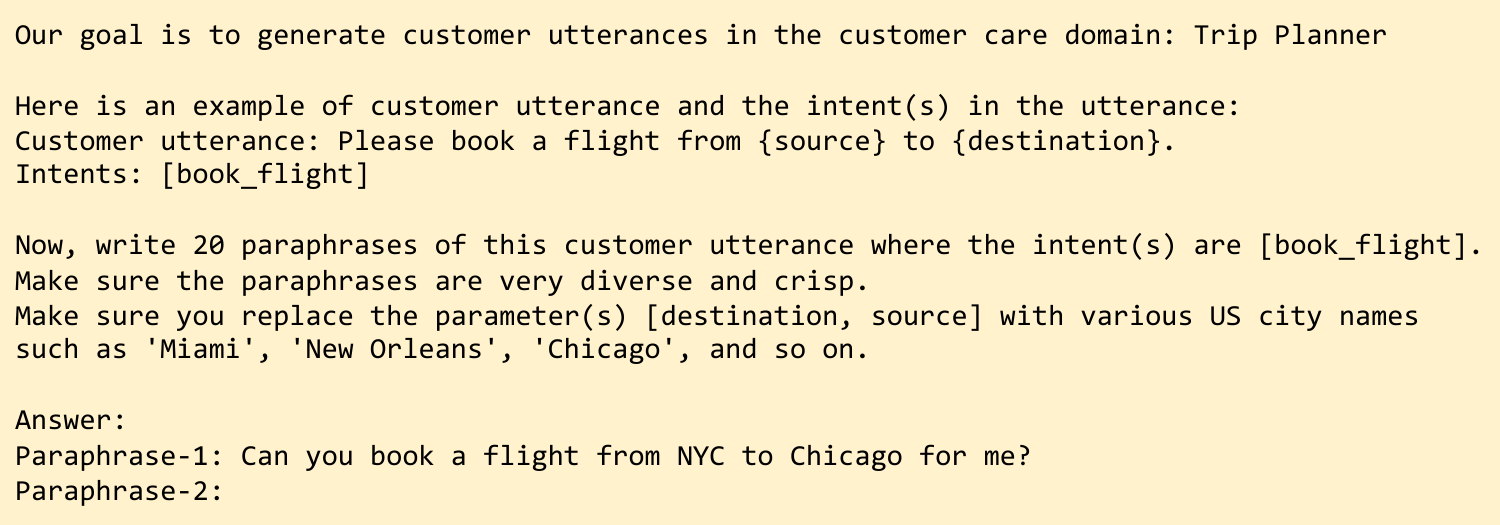}
  \caption{Prompt for generating test utterances for the intent ``book flight'' in the domain ``Trip Booking''.} 
  \label{fig:prompt-for-utterance-generation}
\end{figure*}

The prompt for generating test queries related to the intent ``book flight'' in the domain ``Trip Booking'' is shown in Fig. \ref{fig:prompt-for-utterance-generation}. Similar prompts were used for the generation of queries for all other intents. Some generated user queries related to various intents are shown in Tab. \ref{tab:generated-examples}.

\begin{table*}[t]
\centering
\resizebox{2\columnwidth}{!}{%
\begin{tabular}
{>{\arraybackslash}m{20cm}}
\toprule
\textbf{Intents and Generated Queries in Domain Trip Booking} \\
\midrule
\textbf{Book Car:} Could you arrange for car rental for my Minneapolis trip from June 25th to July 3rd?\\
\textbf{Book Hotel:} I'm looking to book a hotel room in Los Angeles from August 18th to August 25th.\\
\textbf{Book Flight:} I need to fly from Miami to Toronto, can you please help me with that?\\

\midrule
\textbf{Intents and Generated Queries in Domain Insurance}\\
\midrule
\textbf{Buy Insurance:} I want to safeguard my family's health with an insurance policy.\\
\textbf{Cancel Insurance:} My current policy doesn't cover all the expenses, would you please cancel it?\\
\textbf{Add Member:} Would you please add my aunt to my mobile insurance policy?\\

\midrule
\textbf{Intents and Generated Queries in Domain Banking} \\
\midrule
\textbf{Open Account:} My wife and I are looking to open a joint bank account together. What steps do we need to take?\\
\textbf{Report Problem:}  My card is declined at the ATM. Can you help?\\
\textbf{Cancel Transaction:} The balance of my savings account has been altered by an unauthorized transfer of \$500. I would appreciate if you could get that money refunded.\\

\midrule
\textbf{Intents and Generated Queries in Domain Restaurant \& Ride Book} \\
\midrule
\textbf{Book Restaurant:} Could you assist me in making a reservation at a restaurant in Manhattan?\\
\textbf{Book Ride:} I am in Washington D.C. looking for a ride to Fort Lauderdale.\\
\textbf{Cancel Restaurant:} I made a restaurant booking in Los Angeles this morning but I cannot make it due to an emergency. Can you cancel my booking?\\
\textbf{Cancel Ride:} May I cancel my ride from Brooklyn to Queens?\\

\bottomrule
\end{tabular}}
\caption{Examples of generated user queries for each intent.}
\label{tab:generated-examples}
\end{table*}

\section{Prompt Structures for Generating Plans}
\label{sec:appx-prompt-structure-for-plan-generation}
\begin{figure*}[h]
  \centering
  \includegraphics[width=0.8\textwidth]{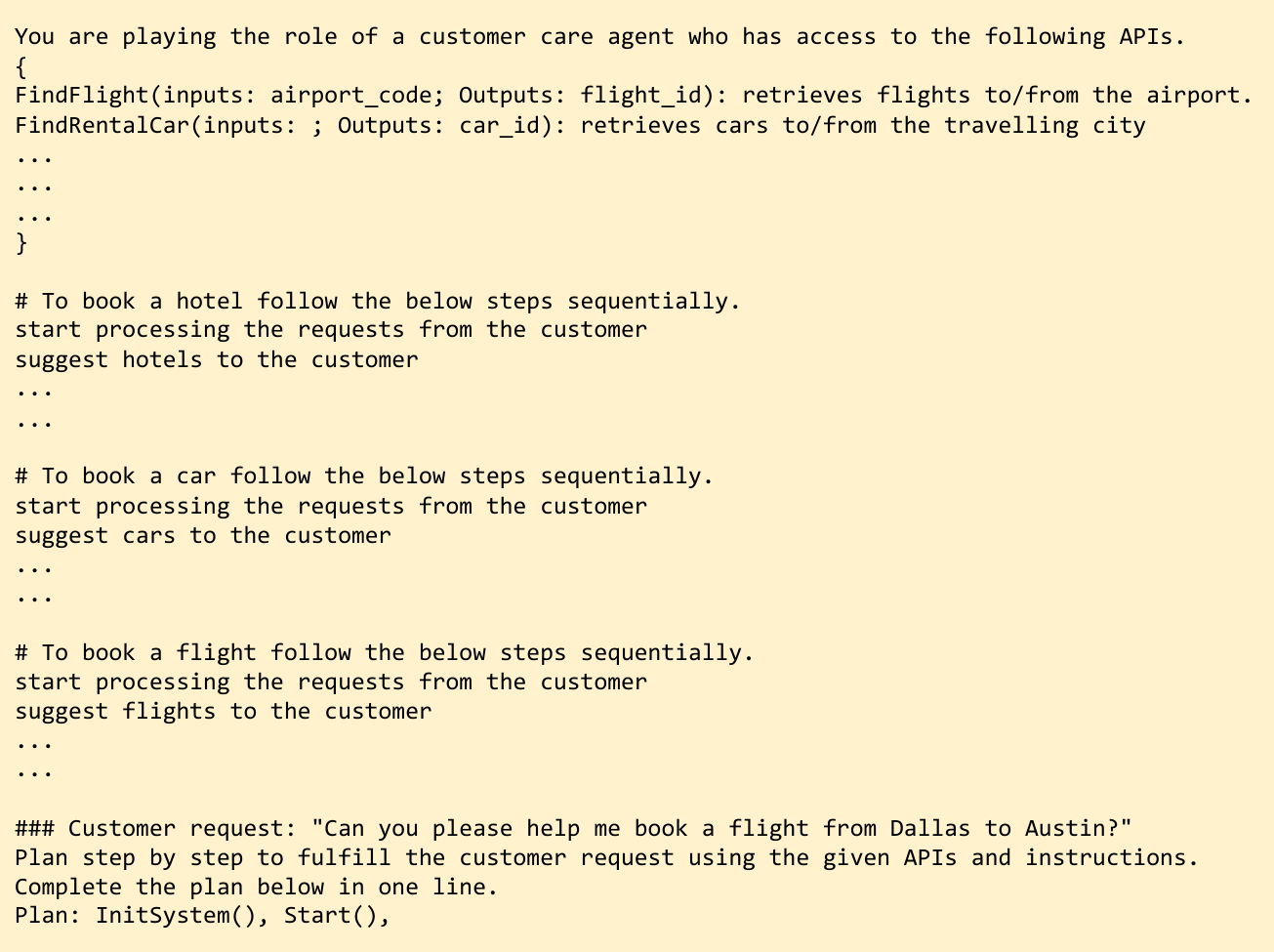}
  \caption{Prompt format for generating a plan consisting of APIs only (without thoughts) in the domain ``Trip Booking''. The first two dummy APIs \texttt{InitSystem(), Start()} are provided in context to guide the model to follow the plan format.} 
  \label{fig:prompt-for-plan-generation-api-only}
\end{figure*}

\begin{figure*}[h]
  \centering
  \includegraphics[width=0.8\textwidth]{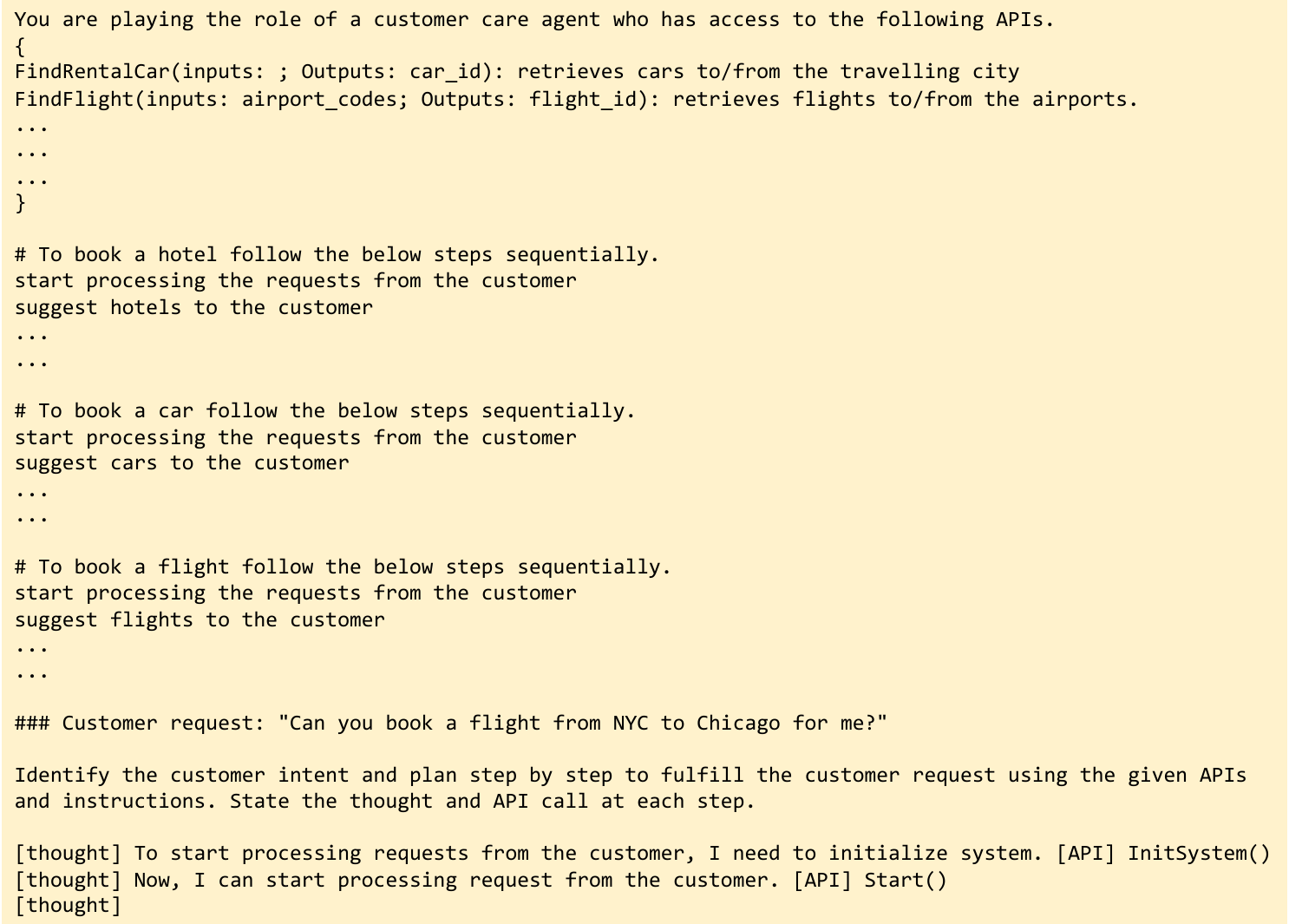}
  \caption{Prompt format for generating a plan (with thoughts) in the domain ``Trip Booking''. The first two dummy steps, e.g., thoughts related to \texttt{InitSystem()} and \texttt{Start()} are provided in context to guide the model to follow the plan format.} 
  \label{fig:prompt-for-plan-generation}
\end{figure*}

The prompt structure for generating plans consisting of only APIs (no thoughts) is shown in Fig. \ref{fig:prompt-for-plan-generation-api-only}. The prompt structure for planning with thoughts and APIs is shown in Fig. \ref{fig:prompt-for-plan-generation}.

\section{Experimental Details}
\label{sec:appx-experimental-details}

\subsection{Hyper-parameter Tuning}
\label{sec:appx-hyperparameter-tuning}
In this section, we discuss the hyper-parameter selection process in FLAP, our proposed constrained decoding based algorithm in \S\ref{sec:proposed-algorithm}.

In our dataset, all user queries are single intent, hence, we set $\alpha_{c}>\alpha_{a}>\alpha_{b}$ ($\alpha_{a}=0.5$, $\alpha_{b}=0.1$, and $\alpha_{c}=1$). We empirically determine $\beta=0.1$ when the APIs are hallucinated or violates the dependency, i.e., a soft alignment of generated APIs to permitted APIs. 

To tune the values of the weight on the heuristic score ($\lambda$) and beam size ($k$), we run an ablation with a subset of the data ($5$ user utterances per intent) in Tab. \ref{tab:domain-and-flow-statistics}, in the setting where only relevant subflow is given in-context, using \texttt{Mistral-7b-instruct}. We experiment with $k=[5, 10, 15]$ and $\lambda=[0.3, 0.5, 0.7]$. The results are summarized in Tab. \ref{tab:hyperparameter-tuning}. We find that $k=10$ and $\lambda=0.7$ result in overall best performance, especially in terms of API dependency violation (refer to ``per plan avg \% of inconsistent APIs''). Hence, we report the results with these values.

We set the values of all the scaling factors ($a$, $b$, $c$, $d$) to $1$ implying equal weight to all heurictic score components. We set the lookahead length, $L=32$ in our experiments. During plan generation, if the API \texttt{Finish()} is generated, we consider that as end-of-plan (as per the flows in Tab. \ref{tab:appx-domain-and-flow-example}) and terminate the generation.

For the Beam Search baseline, we set number of beams to $3$ and use \texttt{no\_repeat\_ngram\_size=10} to prevent the model from generating the same thought and API pair repeatedly. For the Nucleus Sampling baseline we use \texttt{top\_p=0.9}.

\subsection{Infrastructure and Libraries}
\label{sec:appx-infrastructure-and-libraries}
We use eight NVIDIA A10G Tensor Core GPUs for all of our experiments. We use Pytorch\footnote{\url{https://pytorch.org}} for all implementations. For all the LLMs used in this paper, we use their Huggingface\footnote{\url{https://huggingface.co}} implementations. We build our proposed constrained decoding algorithm on top of Huggingface generation module\footnote{\url{https://huggingface.co/docs/transformers/main\_classes/text\_generation}}. We used the Huggingface implementation of the \texttt{all-MiniLM-L6-v2}\footnote{\url{https://huggingface.co/sentence-transformers/all-MiniLM-L6-v2}} sentence transformer for measuring semantic similarities (\texttt{sim()}).

\subsection{Runtime Analysis}
\label{sec:appx-runtime-analysis}
The runtimes of different models in plan generation are shown in Tab. \ref{tab:runtime-analysis}. Note that our proposed constrained decoding algorithm is run on only one GPU of 24GB memory with a batch size of $1$. This is one of the reasons of the high runtime. We explored parallelization techniques by deploying the LLMs in multiple GPUs to make the process faster. However, it was observed that deploying the LLMs (Huggingface implementation) in multiple GPUs made the inference time of the base LLMs even slower. Our intended future work includes more efficient deployment of the LLMs in potentially multi-GPU machines which is out-of-scope of this paper.     

\begin{table}[t]
\centering
\resizebox{1\columnwidth}{!}{%
\begin{tabular}
{>{\centering\arraybackslash}m{0.8cm}|>{\arraybackslash}m{8cm}|>{\centering\arraybackslash}m{1.5cm}|>{\centering\arraybackslash}m{3.5cm}|>{\centering\arraybackslash}m{4cm}}
\toprule
\textbf{LLM} & \textbf{Models} & \textbf{With} & \multicolumn{2}{c}{\textbf{Avgerage Runtime Per Plan (seconds)}}\\

& & \textbf{Thought} & \textbf{All flows in context} & \textbf{Relevant flow in context}\\
\midrule

\multirow{8}{*}{\textbf{\rotatebox[origin=c]{90}{\texttt{mpt-7b-ins.}}}}  &  Greedy Search & No & $7.94$ & $5.4$\\
                                                    &  Greedy Search & Yes & $14.07$ & $11.3$ \\
                                                    &  Beam Search & Yes  & $27.94$ & $26.69$ \\
                                                    &  Nucleus Sampling & Yes & $17.74$ & $17.55$ \\
                                                    &  FLAP.1 [soft api + align thought to (api, intent)] & Yes & $264.0$ & $371.41$  \\
                                                    &  FLAP.2 [soft api + align thought to (step, api)]& Yes & $199.28$ & $295.12$ \\
                                                    &  FLAP.3 [soft api + align thought to (step, intent)] & Yes & $378.43$ & $336.4$\\
                                                    &  FLAP.4 [soft api + align thought to (step, api, intent)] & Yes & $216.16$ & $327.79$\\
         
         \midrule
         \multirow{8}{*}{\textbf{\rotatebox[origin=c]{90}{\texttt{mistral-7b-ins.}}}}  &  Greedy Search & No & $19.72$ & $18.81$\\
                                                        &  Greedy Search & Yes  & $8.78$ & $9.91$\\
                                                        &  Beam Search & Yes & $10.24$ & $9.76$ \\
                                                        &  Nucleus Sampling & Yes & $13.09$ & $11.57$ \\
                                                        & FLAP.1 [soft api + align thought to (api, intent)] & Yes & $250.67$ & $252.45$\\
                                                        & FLAP.2 [soft api + align thought to (step, api)] & Yes & $221.63$ & $234.47$\\
                                                        & FLAP.3 [soft api + align thought to (step, intent)] & Yes & $228.61$ & $117.58$\\
                                                        & FLAP.4 [soft api + align thought to (step, api, intent)] & Yes & $238.82$ & $113.22$\\

\bottomrule

\end{tabular}}
\caption{Average runtime of various models over all generated plans.}
\label{tab:runtime-analysis}
\end{table}

\begin{table*}[t]
\centering

\resizebox{2\columnwidth}{!}{%
\begin{tabular}
{>{\arraybackslash}m{1cm}>{\arraybackslash}m{2cm}>{\arraybackslash}m{2.5cm}>{\centering\arraybackslash}m{1.5cm}|>{\arraybackslash}m{2.2cm}>{\arraybackslash}m{2.2cm}|>{\centering\arraybackslash}m{2cm}|>{\arraybackslash}m{2.4cm}>{\arraybackslash}m{2.6cm}|>{\arraybackslash}m{2.4cm}>{\arraybackslash}m{2.4cm}|>{\arraybackslash}m{2.4cm}>{\arraybackslash}m{2.4cm}}
\toprule

\textbf{LLM}    & \textbf{Varying} & \textbf{Parameter} & \textbf{With}	& \multicolumn{2}{c|}{\textbf{Avg count per plan}} & \textbf{\% Plan}	& \multicolumn{2}{c|}{\textbf{Avg. \% of APIs in plan that are}} & \multicolumn{2}{c|}{\textbf{Avg \# of edit for}}	& \multicolumn{2}{c}{\textbf{Per plan avg \% of inconsistent}}	\\

& \textbf{Parameter} & \textbf{Values} & \textbf{Thoughts}	& \textbf{Thought}	& \textbf{APIs}	& \textbf{Parsable ($\uparrow$)} & \textbf{Repeated ($\downarrow$)}	& \textbf{Hallucinated ($\downarrow$)}	& \textbf{Steps ($\downarrow$)}	& \textbf{APIs ($\downarrow$)}	& \textbf{Steps ($\downarrow$)}	& \textbf{APIs ($\downarrow$)}\\
\midrule
         
         \multirow{7}{*}{\textbf{\rotatebox[origin=c]{90}{\texttt{mistral-7B-Ins.}}}}  &   $\lambda$   &  top k=10, $\lambda$=0.7  & Yes &	$6.85\pm2.96$ &	$6.98\pm3.04$ &	$100.0$ &	$0.81\pm4.65$ &	$0.0\pm2.15$ &	$2.2\pm2.94$ &	$2.19\pm2.8$ &	$6.92\pm9.12$ &	$2.99\pm8.39$\\
                                                        &   $\lambda$  &  top k=10, $\lambda$=0.5  & Yes &	$6.02\pm2.73$ &	$6.03\pm2.72$ &	$100.0$ &	$0.0\pm0.0$ &	$0.0\pm2.46$ &	$2.51\pm2.58$ &	$2.55\pm2.63$ &	$7.06\pm11.59$ &	$16.39\pm15.81$\\
                                                        &   $\lambda$   &  top k=10, $\lambda$=0.3  & Yes &	$5.11\pm1.58$ &	$5.11\pm1.58$ &	$100.0$ &	$0.22\pm1.76$ &	$0.0\pm0.0$ &	$2.09\pm2.01$ &	$2.17\pm1.98$ &	$4.22\pm8.77$ &	$20.62\pm18.75$\\
                                                        & $\lambda$ & $\lambda$=0 (Greedy) & Yes &	$5.66\pm3.48$ &	$5.77\pm3.6$ &	$100.0$ &	$3.07\pm10.67$ &	$1.0\pm4.3$ &	$2.78\pm3.53$ &	$2.86\pm3.45$ &	$3.15\pm7.39$ &	$31.25\pm22.82$\\

\cmidrule{2-13}
                                                        &   top k   &  top k=5, $\lambda$=0.7   & Yes &	$6.52\pm2.25$ &	$6.52\pm2.25$ &	$100.0$ &	$0.0\pm0.0$ &	$0.0\pm2.07$ &	$2.36\pm2.58$ &	$2.42\pm2.63$ &	$7.42\pm9.53$ &	$4.36\pm9.33$\\
                                                        &   top k   &  top k=10, $\lambda$=0.7  & Yes &	$6.85\pm2.96$ &	$6.98\pm3.04$ &	$100.0$ &	$0.81\pm4.65$ &	$0.0\pm2.15$ &	$2.2\pm2.94$ &	$2.19\pm2.8$ &	$6.92\pm9.12$ &	$2.99\pm8.39$\\
                                                        &   top k   &  top k=15, $\lambda$=0.7  & Yes &	$6.02\pm2.11$ &	$6.19\pm2.19$ &	$100.0$ &	$0.31\pm2.25$ &	$0.0\pm2.25$ &	$2.43\pm2.46$ &	$2.46\pm2.48$ &	$10.42\pm11.14$ &	$4.14\pm11.22$\\

\bottomrule
\end{tabular}}
\caption{Ablation for tuning constrained decoding parameters of FLAP: beam size for lookahead (top k) and lookahead heuristic score weight ($\lambda$). The ablation was done using a subset of the data (5 user utterances per intent). There are $6.84$ average APIs per gold plan (reference for the ``Avg. count per plan'' column).}

\label{tab:hyperparameter-tuning}
\end{table*}

\section{Additional Results}
\label{sec:appx-additional-results}
The ablation results with standard deviations over all test data points with the prompting-based plan generation using numerous open-source models in the settings where all flows are given in-context and only relevant flows are given in-context, can be found in Tab. \ref{tab:ablation-all-flows} and Tab. \ref{tab:ablation-relevant-flows}, respectively.

The results with standard deviations for plan generation using our proposed constrained decoding algorithm, FLAP, in the setting where all flows are given in-context and only relevant flows are given in-context, can be found in Tab. \ref{tab:cd-results-all-flows} and Tab. \ref{tab:cd-results-relevant-flows}, respectively.

\begin{table*}[t]
\centering

\resizebox{2\columnwidth}{!}{%
\begin{tabular}
{>{\arraybackslash}m{3.7cm}>{\centering\arraybackslash}m{1.5cm}|>{\arraybackslash}m{2.2cm}>{\arraybackslash}m{2.2cm}|>{\centering\arraybackslash}m{2cm}|>{\arraybackslash}m{2.4cm}>{\arraybackslash}m{2.6cm}|>{\arraybackslash}m{2.4cm}>{\arraybackslash}m{2.4cm}|>{\arraybackslash}m{2.4cm}>{\arraybackslash}m{2.4cm}}
\toprule

\textbf{LLMs}	& \textbf{With}	& \multicolumn{2}{c|}{\textbf{Avg count per plan}}	& \textbf{\% Plan} & \multicolumn{2}{c|}{\textbf{Avg \% of APIs in plan that are}} & \multicolumn{2}{c|}{\textbf{Avg \# of edit for}}	& \multicolumn{2}{c}{\textbf{Per plan avg \% of inconsistent}}	\\
& \textbf{Thought}	& \textbf{Thoughts}	& \textbf{APIs} & \textbf{Parsable ($\uparrow$)}	& \textbf{Repeated ($\downarrow$)}	& \textbf{Hallucinated ($\downarrow$)}	& \textbf{Steps ($\downarrow$)}	& \textbf{APIs ($\downarrow$)}	& \textbf{Steps ($\downarrow$)}	& \textbf{APIs ($\downarrow$)}\\
\midrule
\multirow{2}{*}{\textbf{\texttt{santacoder-1.1b}}}	& No	& -                   & $33.73\pm49.05$ & $100.0$	& $19.84\pm37.01$	& $1.21\pm9.43$     & $30.34\pm47.63$	& $25.91\pm39.8$	& $15.42\pm14.02$	& $33.06\pm24.21$\\
                                        & Yes	& $15.09\pm8.66$      & $14.98\pm8.59$ & $99.62$	& $27.61\pm30.64$	& $13.0\pm21.75$	& $11.78\pm7.66$	& $13.02\pm8.27$	& $18.17\pm16.39$	& $34.13\pm27.51$\\
\midrule

\multirow{2}{*}{\textbf{\texttt{toolAlpaca-7b}}}	& No & 	- & 	$4.13\pm2.16$ & 	$100.0$ &	$0.0\pm0.0$ &	$10.76\pm23.62$ &	$4.64\pm2.05$ &	$5.12\pm2.42$ &	$20.41\pm19.98$ &	$49.6\pm26.21$\\
                                                & Yes &	$4.88\pm4.26$ &	$4.88\pm4.26$ &	$90.38$ &	$3.32\pm12.31$ &	$19.0\pm21.98$ &	$4.67\pm4.07$ &	$5.3\pm3.23$ &	$17.6\pm17.27$ &	$44.78\pm25.18$\\
\midrule

\multirow{2}{*}{\textbf{\texttt{falcon-7b-instruct}}}	& No	& -               & $55.97\pm56.1$ & $100.0$	& $42.65\pm44.96$	& $6.06\pm19.0$	& $47.48\pm51.98$	& $40.34\pm45.91$	& $16.61\pm20.56$	& $27.67\pm26.48$\\
                                                & Yes	& $11.62\pm6.44$  & $11.69\pm6.31$	& $100.0$ & $39.02\pm30.8$	& $19.0\pm27.34$	& $9.73\pm6.45$	& $11.41\pm7.34$	& $24.03\pm25.24$	& $38.83\pm26.52$\\
\midrule
\multirow{2}{*}{\textbf{\texttt{mpt-7b-instruct}}}	&	No &	- &	$17.32\pm32.42$ &	$100.0$ &	$11.58\pm28.7$ &	$0.17\pm1.39$ &	$16.17\pm31.7$ &	$13.42\pm24.88$ &	$14.65\pm14.12$ &	$33.33\pm21.74$\\
                                            &	Yes &	$16.2\pm7.59$ &	$16.08\pm7.49$ &	$97.69$ &	$35.11\pm30.54$ &	$4.0\pm10.64$ &	$13.92\pm8.03$ &	$12.38\pm7.69$ &	$11.86\pm15.74$ &	$29.61\pm23.69$\\
\midrule
\multirow{2}{*}{\textbf{\texttt{mistral-7b-instruct}}}	&	No &	-	& $4.78\pm1.7$ &	$100.0$ &	$0.32\pm3.04$ &	$2.36\pm7.04$ &	$2.87\pm1.75$ &	$3.02\pm1.79$ &	$10.6\pm15.84$ &	$37.23\pm25.28$\\
                                                &	Yes &	$4.72\pm2.38$ &	$4.76\pm2.41$ &	$99.62$ &	$2.92\pm11.73$ &	$0.0\pm3.59$ &	$2.78\pm2.56$ &	$2.8\pm2.53$ &	$3.09\pm8.21$ &	$40.31\pm23.89$\\
\midrule
\multirow{2}{*}{\textbf{\texttt{koala-13b}}}	& No	& -	               & $45.79\pm44.81$ & $100.0$	& $39.64\pm41.95$	& $6.29\pm17.54$	& $37.6\pm40.87$	& $31.4\pm32.87$	& $10.51\pm13.37$	& $23.98\pm18.54$\\
                                    & Yes	& $12.06\pm4.15$   & $12.2\pm4.21$ & $97.31$	& $24.37\pm30.11$	& $10.0\pm20.04$	& $8.76\pm5.23$	& $9.08\pm5.52$	& $9.02\pm11.68$	& $24.04\pm19.39$\\
\midrule
\multirow{2}{*}{\textbf{\texttt{vicuna-13b}}}	& No	& -	            & $5.94\pm2.68$	& $100.0$ & $0.57\pm2.75$	& $2.38\pm6.99$	& $4.12\pm2.51$	& $4.29\pm2.54$	& $15.08\pm16.23$	& $35.22\pm20.51$\\
                                        & Yes	& $6.94\pm3.53$	& $6.99\pm3.6$ & $92.31$	& $4.46\pm9.76$	& $4.0\pm9.55$	& $4.02\pm3.03$	& $4.28\pm3.11$	& $6.73\pm11.49$	& $35.49\pm22.61$\\
\midrule
\multirow{2}{*}{\textbf{\texttt{llama-13b}}}	& No	& -	& $29.41\pm37.77$ & $100.0$	& $25.84\pm38.81$	& $2.36\pm9.19$	& $26.77\pm37.17$	& $20.87\pm29.83$	& $13.16\pm16.08$	& $31.72\pm25.31$\\
                                    & Yes	& $10.21\pm6.9$	& $10.13\pm6.79$ & $98.85$	& $21.87\pm33.24$	& $4.0\pm12.7$	& $9.18\pm7.12$	& $8.66\pm7.05$	& $4.68\pm10.61$	& $31.04\pm27.63$\\
\midrule
\multirow{2}{*}{\textbf{\texttt{mpt-30b-chat}}}	& No	& -	& $5.7\pm1.93$ & $100.0$	& $1.73\pm5.81$	& $0.53\pm3.94$	& $3.26\pm2.13$	& $3.35\pm2.08$	& $9.08\pm14.36$	& $33.41\pm19.78$\\
                                        & Yes	& $5.57\pm2.07$	& $5.87\pm2.25$ & $96.92$	& $2.91\pm7.67$	& $1.0\pm5.15$	& $2.7\pm2.58$	& $2.7\pm2.6$	& $5.72\pm10.18$	& $32.45\pm22.35$\\
\midrule
\multirow{2}{*}{\textbf{\texttt{falcon-40b-instruct}}}	& No	& -	& $10.21\pm21.88$ & $100.0$	& $3.1\pm15.86$	& $0.89\pm4.24$	& $8.07\pm19.65$	& $7.15\pm14.25$	& $18.0\pm17.61$	& $38.27\pm20.27$\\
                                                & Yes	& $7.6\pm4.83$	& $7.75\pm4.96$ & $100.0$	& $10.25\pm21.52$	& $5.0\pm12.61$	& $5.35\pm4.55$	& $5.4\pm4.29$	& $8.95\pm12.34$	& $36.42\pm24.54$\\

\bottomrule

\end{tabular}}
\caption{Zero-shot plan generation results using greedy decoding when all flows are given in the prompt. There are $6.84$ average APIs per gold plan (reference for the ``Avg count per plan'' column).}

\label{tab:ablation-all-flows}
\end{table*}

\begin{table*}[t]
\centering

\resizebox{2\columnwidth}{!}{%
\begin{tabular}
{>{\arraybackslash}m{3.7cm}>{\centering\arraybackslash}m{1.5cm}|>{\arraybackslash}m{2.2cm}>{\arraybackslash}m{2.2cm}|>{\centering\arraybackslash}m{2cm}|>{\arraybackslash}m{2.4cm}>{\arraybackslash}m{2.6cm}|>{\arraybackslash}m{2.4cm}>{\arraybackslash}m{2.4cm}|>{\arraybackslash}m{2.4cm}>{\arraybackslash}m{2.4cm}}
\toprule

\textbf{LLMs}	& \textbf{With}	& \multicolumn{2}{c|}{\textbf{Avg count per plan}} & \textbf{\% Plan}	& \multicolumn{2}{c|}{\textbf{Avg \% of APIs in plan that are}} & \multicolumn{2}{c|}{\textbf{Avg \# of edit for}}	& \multicolumn{2}{c}{\textbf{Per plan avg \% of inconsistent}}	\\
& \textbf{Thought}	& \textbf{Thoughts}	& \textbf{APIs} & \textbf{Parsable ($\uparrow$)}	& \textbf{Repeated ($\downarrow$)}	& \textbf{Hallucinated ($\downarrow$)}	& \textbf{Steps ($\downarrow$)}	& \textbf{APIs ($\downarrow$)}	& \textbf{Steps ($\downarrow$)}	& \textbf{APIs ($\downarrow$)}\\
\midrule
         \multirow{2}{*}{\textbf{\texttt{santacoder-1.1b}}}  &  No&  -&  $34.08\pm49.7$ & $100.0$ &  $19.81\pm37.22$&  $1.79\pm11.73$&  $29.58\pm46.88$&  $25.75\pm39.29$&  $15.34\pm15.5$& $34.14\pm23.77$\\
                                                    &  Yes&  $13.14\pm9.84$&  $13.07\pm9.76$ & $100.0$ &  $22.27\pm29.81$&  $20.0\pm28.74$&  $9.48\pm8.03$&  $11.53\pm8.73$&  $19.48\pm19.06$& $31.21\pm25.47$\\
\midrule

\multirow{2}{*}{\textbf{\texttt{toolAlpaca-7b}}}	& No &	- &	$4.13\pm1.88$ &	$100.0$ &	$0.0\pm0.0$ &	$6.99\pm18.96$ &	$4.38\pm2.25$ &	$4.73\pm2.52$ &	$17.77\pm19.97$ &	$49.17\pm26.49$\\
                                                & Yes &	$3.69\pm1.01$ &	$3.7\pm1.02$ &	$96.15$ &	$2.19\pm7.56$ &	$8.0\pm16.92$ &	$3.01\pm1.82$ &	$3.52\pm2.27$ &	$10.56\pm15.83$ &	$47.59\pm22.9$\\

\midrule
         \multirow{2}{*}{\textbf{\texttt{falcon-7b-instruct}}}   &  No&  -&  $50.45\pm52.86$& $100.0$ &  $39.54\pm45.13$&  $6.63\pm20.89$&  $40.76\pm48.73$&  $38.09\pm44.17$&  $17.43\pm20.33$& $23.85\pm21.46$\\
                                                        &  Yes&  $11.52\pm8.03$&  $11.47\pm7.95$& $97.69$ &  $37.26\pm33.21$&  $9.0\pm19.72$&  $9.74\pm8.56$&  $9.15\pm7.74$&  $21.49\pm24.2$& $37.91\pm28.25$\\
\midrule
         \multirow{2}{*}{\textbf{\texttt{mpt-7b-instruct}}}  & No &	- &	$14.89\pm29.87$&	$100.0$ &	$8.5\pm25.06$&	$0.19\pm1.83$&	$13.55\pm29.35$&	$11.52\pm24.08$&	$13.52\pm14.29$&	$35.64\pm20.75$\\
                                                    &	Yes&	$12.42\pm8.3$&	$12.34\pm8.19$&	$98.85$&	$23.02\pm30.15$&	$3.0\pm9.44$&	$9.87\pm8.32$&	$8.93\pm7.8$&	$9.09\pm15.46$&	$33.44\pm23.48$\\
\midrule
         \multirow{2}{*}{\textbf{\texttt{mistral-7b-instruct}}}  &	No&	-&	$5.85\pm2.61$&	$100.0$&	$0.57\pm6.06$&	$1.38\pm5.06$&	$3.1\pm2.93$&	$3.13\pm2.14$&	$8.02\pm12.87$&	$30.61\pm20.76$\\
                                                        &	Yes&	$5.57\pm3.1$&	$5.7\pm3.5$&	$100.0$&	$3.76\pm12.18$&	$1.0\pm3.97$&	$2.65\pm3.48$&	$2.6\pm2.85$&	$2.75\pm6.78$&	$34.36\pm22.77$\\
\midrule

         \multirow{2}{*}{\textbf{\texttt{koala-13b}}}    &  No&  -&  $36.08\pm41.04$& $100.0$ &  $28.75\pm39.8$&  $4.18\pm13.41$&  $29.37\pm37.85$&  $23.78\pm29.36$&  $11.25\pm13.33$& $26.15\pm17.19$\\
                                                & Yes& $10.71\pm4.44$& $10.99\pm4.78$& $100.0$ & $20.25\pm28.69$& $9.0\pm20.52$& $7.42\pm5.61$& $7.93\pm6.06$& $9.28\pm10.64$&$22.64\pm17.01$\\
\midrule                                                
         \multirow{2}{*}{\textbf{\texttt{vicuna-13b}}}   & No& -& $5.98\pm2.53$& $100.0$ & $0.32\pm1.95$& $2.4\pm7.3$& $3.91\pm2.51$& $4.12\pm2.5$& $14.21\pm15.44$&$31.66\pm17.25$\\
                                                & Yes& $6.12\pm3.08$& $6.19\pm3.14$& $92.31$ & $3.68\pm8.51$& $2.0\pm7.09$& $3.08\pm2.68$& $3.37\pm2.79$& $5.4\pm9.27$&$35.02\pm22.05$\\
\midrule
 
         \multirow{2}{*}{\textbf{\texttt{llama-13b}}}    & No& -& $27.92\pm38.12$& $100.0$ & $23.04\pm38.37$& $2.0\pm9.53$& $24.68\pm37.43$& $19.1\pm30.62$& $11.34\pm14.81$&$29.43\pm22.09$\\
                                                & Yes& $8.73\pm6.97$& $8.67\pm6.86$& $100.0$ & $18.87\pm33.3$& $3.0\pm10.05$& $6.58\pm7.58$& $5.6\pm6.28$& $4.03\pm12.27$&$38.39\pm30.38$\\
\midrule
         \multirow{2}{*}{\textbf{\texttt{mpt-30b-chat}}} & No& -& $6.68\pm5.29$& $100.0$ & $3.88\pm11.23$& $0.19\pm1.78$& $3.53\pm5.38$& $3.33\pm3.05$& $6.32\pm10.9$&$33.33\pm20.31$\\
                                                & Yes& $5.43\pm2.22$& $5.73\pm2.32$& $97.31$ & $3.12\pm8.4$& $1.0\pm3.97$& $2.13\pm2.05$& $2.09\pm1.86$& $4.56\pm9.25$&$34.33\pm23.63$\\
\midrule
         \multirow{2}{*}{\textbf{\texttt{falcon-40b-instruct}}}  & No& -& $11.22\pm23.02$& $100.0$ & $4.01\pm18.09$& $0.89\pm4.63$& $8.97\pm21.72$& $7.87\pm16.12$& $17.88\pm18.08$&$35.5\pm20.26$\\
                                                        & Yes& $7.14\pm4.31$& $7.26\pm4.37$& $97.69$ & $10.55\pm21.04$& $7.00\pm14.17$& $3.95\pm4.3$& $4.11\pm4.1$& $8.66\pm15.45$&$25.68\pm20.45$\\
    \bottomrule

\end{tabular}}
\caption{Zero-shot plan generation results using greedy decoding when only the flow related to the user query is given in the prompt. There are $6.84$ average APIs per gold plan (reference for the ``Avg count per plan'' column).}

\label{tab:ablation-relevant-flows}
\end{table*}

\begin{table*}[t]
\centering
\resizebox{2\columnwidth}{!}{%
\begin{tabular}
{>{\centering\arraybackslash}m{1cm}|>{\arraybackslash}m{8cm}>{\centering\arraybackslash}m{1.5cm}|>{\arraybackslash}m{2.2cm}>{\arraybackslash}m{2.2cm}|>{\centering\arraybackslash}m{2cm}|>{\arraybackslash}m{2.4cm}>{\arraybackslash}m{2.6cm}|>{\arraybackslash}m{2.4cm}>{\arraybackslash}m{2.4cm}|>{\arraybackslash}m{2.4cm}>{\arraybackslash}m{2.4cm}}
\toprule

\textbf{LLMs}    & \textbf{Decoding Strategy} & \textbf{With}	& \multicolumn{2}{c|}{\textbf{Avg count per plan}}	& \textbf{\% Plan} & \multicolumn{2}{c|}{\textbf{Avg \% of APIs in plan that are}} & \multicolumn{2}{c|}{\textbf{Avg \# of edit for}}	& \multicolumn{2}{c}{\textbf{Per plan avg \% of inconsistent}}	\\

& & \textbf{Thought}	& \textbf{Thoughts}	& \textbf{APIs}	& \textbf{Parsable ($\uparrow$)} & \textbf{Repeated ($\downarrow$)}	& \textbf{Hallucinated ($\downarrow$)}	& \textbf{Steps ($\downarrow$)}	& \textbf{APIs ($\downarrow$)}	& \textbf{Steps ($\downarrow$)}	& \textbf{APIs ($\downarrow$)}\\
\midrule
         \multirow{8}{*}{\textbf{\rotatebox[origin=c]{90}{\texttt{mpt-7b-ins.}}}}  &  Greedy Search & No &	- &	$17.32\pm32.42$ &	$100.0$ &	$11.58\pm28.7$ &	$0.17\pm1.39$ &	$16.17\pm31.7$ &	$13.42\pm24.88$ &	$14.65\pm14.12$ &	$33.33\pm21.74$\\
                                                    &  Greedy Search & Yes &	$16.2\pm7.59$ &	$16.08\pm7.49$ &	$97.69$ &	$35.11\pm30.54$ &	$4.0\pm10.64$ &	$13.92\pm8.03$ &	$12.38\pm7.69$ &	$11.86\pm15.74$ &	$29.61\pm23.69$\\

                                                    & Beam Search &	Yes &	$9.94\pm5.82$ &	$9.9\pm5.8$ &	$70.77$ &	$5.48\pm11.7$ &	$3.0\pm7.31$ &	$7.56\pm4.6$ &	$7.73\pm4.75$ &	$13.93\pm14.06$ &	$40.99\pm19.49$\\
                                                    & Nucleus Sampling &	Yes &	$12.3\pm6.4$ &	$12.41\pm6.4$ &	$76.92$ &	$25.28\pm22.63$ &	$4.0\pm12.28$ &	$9.84\pm5.73$ &	$9.46\pm6.06$ &	$13.82\pm15.15$ &	$31.72\pm18.74$\\

                                                    \cmidrule{2-12}
                                                    
                                                    &  FLAP.1 [soft api + align thought to (api, intent)]& Yes &	$10.25\pm3.63$ &	$10.37\pm3.68$ &	$100.0$ &	$1.26\pm6.03$ &	$0.0\pm0.36$ &	$5.51\pm4.14$ &	$5.47\pm4.02$ &	$6.3\pm7.7$ &	$1.26\pm4.28$\\
                                                    &  FLAP.2 [soft api + align thought to (step, api)]& Yes &	$9.28\pm4.54$ &	$9.37\pm4.57$ &	$100.0$ &	$2.99\pm11.26$ &	$0.0\pm0.62$ &	$5.68\pm5.01$ &	$5.58\pm4.85$ &	$4.34\pm7.22$ &	$1.62\pm5.15$\\
                                                    &  FLAP.3 [soft api + align thought to (step, intent)]& Yes &	$9.16\pm4.02$ &	$9.39\pm4.07$ &	$100.0$ &	$1.04\pm6.04$ &	$0.0\pm0.0$ &	$5.02\pm3.92$ &	$4.98\pm3.87$ &	$6.1\pm8.06$ &	$1.32\pm4.34$\\
                                                    &  FLAP.4 [soft api + align thought to (step, api, intent)]& Yes &	$9.29\pm4.5$ &	$9.38\pm4.55$ &	$100.0$ &	$3.63\pm11.94$ &	$0.0\pm1.55$ &	$5.53\pm4.93$ &	$5.41\pm4.74$ &	$4.81\pm7.71$ &	$2.05\pm6.53$\\
         
         \midrule
         \multirow{8}{*}{\textbf{\rotatebox[origin=c]{90}{\texttt{mistral-7b-ins.}}}}  &  Greedy Search & No &	- &	$4.78\pm1.7$ &	$100.0$ &	$0.32\pm3.04$ &	$2.36\pm7.04$ &	$2.87\pm1.75$ &	$3.02\pm1.79$ &	$10.6\pm15.84$ &	$37.23\pm25.28$\\
                                                        &  Greedy Search & Yes &	$4.72\pm2.38$ &	$4.76\pm2.41$ &	$99.62$ &	$2.92\pm11.73$ &	$0.0\pm3.59$ &	$2.78\pm2.56$ &	$2.8\pm2.53$ &	$3.09\pm8.21$ &	$40.31\pm23.89$\\

                                                        & Beam Search &	Yes &	$3.61\pm1.44$ &	$3.75\pm1.62$ &	$98.08$ &	$0.2\pm2.05$ &	$0.0\pm1.55$ &	$2.88\pm2.19$ &	$2.95\pm2.15$ &	$2.37\pm8.37$ &	$38.94\pm26.87$\\
                                                        & Nucleus Sampling &	Yes &	$5.67\pm2.83$ &	$5.91\pm2.87$ &	$93.85$ &	$5.79\pm12.25$ &	$2.0\pm5.77$ &	$2.97\pm2.51$ &	$3.02\pm2.56$ &	$7.47\pm13.13$ &	$34.28\pm22.38$\\

                                                        \cmidrule{2-12}
                                                        
                                                        & FLAP.1 [soft api + align thought to (api, intent)]& Yes &	$6.62\pm2.72$ &	$6.65\pm2.73$ &	$100.0$ &	$0.0\pm0.0$ &	$0.0\pm2.85$ &	$2.45\pm3.18$ &	$2.51\pm3.23$ &	$6.26\pm9.47$ &	$4.59\pm10.35$\\
                                                        & FLAP.2 [soft api + align thought to (step, api)]& Yes &	$6.09\pm2.02$ &	$6.12\pm2.09$ &	$100.0$ &	$0.12\pm1.14$ &	$0.0\pm3.03$ &	$2.25\pm3.08$ &	$2.32\pm3.16$ &	$5.06\pm8.57$ &	$5.47\pm11.81$\\
                                                        & FLAP.3 [soft api + align thought to (step, intent)]& Yes &	$6.27\pm2.21$ &	$6.36\pm2.21$ &	$100.0$ &	$0.39\pm3.59$ &	$0.0\pm0.0$ &	$1.97\pm2.89$ &	$2.01\pm2.84$ &	$6.46\pm9.45$ &	$3.63\pm9.06$\\
                                                        & FLAP.4 [soft api + align thought to (step, api, intent)]& Yes &	$6.37\pm2.45$ &	$6.39\pm2.44$ &	$100.0$ &	$0.54\pm5.12$ &	$0.0\pm2.93$ &	$2.45\pm3.05$ &	$2.49\pm3.04$ &	$5.63\pm9.88$ &	$7.68\pm13.15$\\

\bottomrule

\end{tabular}}
\caption{Zero-shot plan generation results with FLAP and other baselines, when all flows are given in the prompt. Here, the numbers (FLAP.\#) indicate different ablation versions of FLAP. Structural constraint is applied in all versions of FLAP. There are $6.84$ average APIs per gold plan (reference for the ``Avg count per plan'' column).}

\label{tab:cd-results-all-flows}
\end{table*}

\begin{table*}[t]
\centering

\resizebox{2\columnwidth}{!}{%
\begin{tabular}
{>{\centering\arraybackslash}m{1cm}|>{\arraybackslash}m{8cm}>{\centering\arraybackslash}m{1.5cm}|>{\arraybackslash}m{2.2cm}>{\arraybackslash}m{2.2cm}|>{\centering\arraybackslash}m{2cm}|>{\arraybackslash}m{2.4cm}>{\arraybackslash}m{2.6cm}|>{\arraybackslash}m{2.4cm}>{\arraybackslash}m{2.4cm}|>{\arraybackslash}m{2.4cm}>{\arraybackslash}m{2.4cm}}
\toprule

\textbf{LLMs}    & \textbf{Decoding Strategy} & \textbf{With}	& \multicolumn{2}{c|}{\textbf{Avg count per plan}}	& \textbf{\% Plan} & \multicolumn{2}{c|}{\textbf{Avg \% of APIs in plan that are}} & \multicolumn{2}{c|}{\textbf{Avg \# of edit for}}	& \multicolumn{2}{c}{\textbf{Per plan avg \% of inconsistent}}	\\

& & \textbf{Thoughts}	& \textbf{Thoughts}	& \textbf{APIs}	& \textbf{Parsable ($\uparrow$)} & \textbf{Repeated ($\downarrow$)}	& \textbf{Hallucinated ($\downarrow$)}	& \textbf{Steps ($\downarrow$)}	& \textbf{APIs ($\downarrow$)}	& \textbf{Steps ($\downarrow$)}	& \textbf{APIs ($\downarrow$)}\\
\midrule
    \multirow{8}{*}{\textbf{\rotatebox[origin=c]{90}{\texttt{mpt-7b-ins.}}}}  &  Greedy Search & No &	- &	$14.89\pm29.87$ &	$100.0$ &	$8.5\pm25.06$ &	$0.19\pm1.83$ &	$13.55\pm29.35$ &	$11.52\pm24.08$ &	$13.52\pm14.29$ &	$35.64\pm20.75$\\
                                                    &  Greedy Search & Yes &	 $12.42\pm8.3$ &	$12.34\pm8.19$ &	$98.85$ &	$23.02\pm30.15$ &	$3.0\pm9.44$ &	$9.87\pm8.32$ &	$8.93\pm7.8$ &	$9.09\pm15.46$ &	$33.44\pm23.48$\\
                                                    & Beam Search &	Yes &	$7.96\pm5.7$ &	$7.96\pm5.66$ &	$81.54$ &	$4.22\pm10.69$ &	$2.0\pm6.43$ &	$5.18\pm4.41$ &	$5.3\pm4.43$ &	$10.81\pm13.0$ &	$41.44\pm19.33$\\
                                                    & Nucleus Sampling &	Yes &	$9.98\pm5.29$ &	$10.27\pm5.58$ &	$75.0$ &	$19.82\pm20.03$ &	$5.0\pm10.53$ &	$7.19\pm5.2$ &	$7.07\pm4.94$ &	$13.55\pm15.4$ &	$33.01\pm21.27$\\

    \cmidrule{2-12}
                                                    &  FLAP.1 [soft api +  align thought to (api, intent)]& Yes &	$9.55\pm3.83$ &	$9.63\pm3.89$ &	$100.0$ &	$1.1\pm6.69$ &	$0.0\pm1.59$ &	$4.67\pm4.05$ &	$4.65\pm3.89$ &	$6.58\pm8.19$ &	$1.2\pm4.13$\\
                                                    &  FLAP.2 [soft api +  align thought to (step, api)] & Yes &	$8.15\pm3.68$ &	$8.2\pm3.69$ &	$100.0$ &	$0.86\pm5.29$ &	$0.0\pm0.0$ &	$3.46\pm3.99$ &	$3.42\pm3.91$ &	$5.1\pm8.36$ &	$1.24\pm5.05$\\
                                                    &  FLAP.3 [soft api +  align thought to (step, intent)]& Yes &	$8.83\pm4.02$ &	$9.0\pm4.06$ &	$100.0$ &	$1.22\pm7.4$ &	$0.0\pm0.0$ &	$4.22\pm4.07$ &	$4.15\pm3.92$ &	$6.27\pm7.77$ &	$2.41\pm7.78$\\
                                                    &  FLAP.4 [soft api +  align thought to (step, api, intent)]& Yes &	$8.6\pm4.48$ &	$8.65\pm4.49$ &	$100.0$ &	$2.99\pm11.21$ &	$0.0\pm0.0$ &	$3.69\pm4.44$ &	$3.37\pm3.9$ &	$5.61\pm9.66$ &	$3.66\pm9.98$\\
    
    \midrule
    \multirow{8}{*}{\textbf{\rotatebox[origin=c]{90}{\texttt{mistral-7b-ins.}}}}   &  Greedy Search & No &	- &	$5.85\pm2.61$ &	$100.0$ &	$0.57\pm6.06$ &	$1.38\pm5.06$ &	$3.1\pm2.93$ &	$3.13\pm2.14$ &	$8.02\pm12.87$ &	$30.61\pm20.76$\\
                                                    & Greedy Search  & Yes &	$5.57\pm3.1$ &	$5.7\pm3.5$ &	$100.0$ &	$3.76\pm12.18$ &	$1.0\pm3.97$ &	$2.65\pm3.48$ &	$2.6\pm2.85$ &	$2.75\pm6.78$ &	$34.36\pm22.77$\\
                                                    & Beam Search &	Yes &	$4.2\pm1.2$ &	$4.28\pm1.26$ &	$100.0$ &	$0.0\pm0.0$ &	$0.0\pm0.0$ &	$2.43\pm1.96$ &	$2.51\pm1.9$ &	$1.54\pm5.59$ &	$36.72\pm23.34$\\
                                                    & Nucleus Sampling &	Yes &	$5.63\pm2.19$ &	$5.94\pm2.42$ &	$98.08$ &	$3.73\pm9.88$ &	$1.0\pm4.71$ &	$2.94\pm2.44$ &	$2.95\pm2.48$ &	$6.77\pm12.65$ &	$30.2\pm20.14$\\

    \cmidrule{2-12}
                                                    & FLAP.1 [soft api +  align thought to (api, intent)]& Yes &	$6.99\pm2.94$ &	$7.01\pm2.95$ &	$100.0$ &	$0.09\pm1.02$ &	$1.0\pm3.47$ &	$2.68\pm3.35$ &	$2.75\pm3.39$ &	$6.01\pm9.7$ &	$4.06\pm9.32$\\
                                                    & FLAP.2 [soft api +  align thought to (step, api)]& Yes &	$6.61\pm2.33$ &	$6.63\pm2.35$ &	$100.0$ &	$0.44\pm4.76$ &	$1.0\pm3.35$ &	$2.29\pm2.86$ &	$2.32\pm2.86$ &	$5.94\pm10.51$ &	$7.01\pm13.06$\\
                                                    & FLAP.3 [soft api +  align thought to (step, intent)]& Yes &	$6.73\pm2.56$ &	$6.82\pm2.61$ &	$100.0$ &	$0.36\pm2.74$ &	$0.0\pm1.05$ &	$2.27\pm2.63$ &	$2.3\pm2.65$ &	$6.58\pm8.78$ &	$4.29\pm9.98$\\
                                                    & FLAP.4 [soft api +  align thought to (step, api, intent)]& Yes &	$6.36\pm2.32$ &	$6.37\pm2.31$ &	$100.0$ &	$0.37\pm3.06$ &	$1.0\pm4.4$ &	$2.2\pm2.76$ &	$2.23\pm2.73$ &	$6.11\pm10.14$ &	$8.66\pm14.1$\\

\bottomrule

\end{tabular}}
\caption{Zero-shot plan generation results with FLAP and other baselines, when only the flow related to the user query is given in the prompt. Here, the numbers (FLAP.\#) indicate different ablation versions of FLAP. Structural constraint is applied in all versions of FLAP. There are $6.84$ average APIs per gold plan (reference for the ``Avg count per plan'' column).}

\label{tab:cd-results-relevant-flows}
\end{table*}

\section{Qualitative Error Analysis}
\label{sec:appx-qualitative-error-analysis}

\subsection{Model entering into repetitive loop with thoughts}
\label{repetition-error}
Example of error when model enters into a repetitive loop when prompted to generate plans with thoughts, can be observed in Tab. \ref{tab:repetition-error}.

\begin{table}[t]
\centering
\resizebox{1\columnwidth}{!}{%
\begin{tabular}
{>{\arraybackslash}m{13cm}}
{\textbf{Query: I want to go from San Diego to Dallas, can you please help me with that?}}\\
\toprule
\textbf{Generated plan by \texttt{Mpt-7b-instruct} using prompting with greedy decoding}\\
$[$thought$]$ I can suggest flights to the customer. $[$API$]$ FindFlight()\\
$[$thought$]$ I can suggest hotels to the customer. $[$API$]$ FindHotel()\\
$[$thought$]$ I can suggest cars to the customer. $[$API$]$ FindRentalCar()\\
$[$thought$]$ I can suggest insurances to the customer. $[$API$]$ GetCarInsuranceDiscount()\\
$[$thought$]$ I can suggest flights to the customer. $[$API$]$ FindFlight()\\
$[$thought$]$ I can suggest hotels to the customer. $[$API$]$ FindHotel()\\
$[$thought$]$ I can suggest cars to the customer. $[$API$]$ FindRentalCar()\\
$[$thought$]$ I can suggest insurances to the customer. $[$API$]$ GetCarInsuranceDiscount()\\
$[$thought$]$ I can suggest flights to the customer. $[$API$]$ FindFlight()\\
$[$thought$]$ I can suggest hotels to the customer. $[$API$]$ FindHotel()\\
$[$thought$]$ I can suggest cars to the customer. $[$API$]$ FindRentalCar()\\
$[$thought$]$ I can suggest insurances to the customer. $[$API$]$ GetCarInsuranceDiscount()\\
$[$thought$]$ I can suggest flights to the customer. $[$API$]$ FindFlight()\\
$[$thought$]$ I can suggest hotels to the customer. $[$API$]$ FindHotel()\\
$[$thought$]$ I can suggest cars to the customer. $[$API$]$ FindRentalCar()\\
\bottomrule
\end{tabular}}
\caption{Example of error when the LLMs enter into a repetitive loop of thought+API in Greedy decoding when prompted to generate thoughts and APIs.}
\label{tab:repetition-error}
\end{table}

\subsection{API hallucinations by pre-trained LLMs with tool data}
\label{sec:appx-hallucination-error}
API hallucination examples by \texttt{Alpaca-7B} are shown in Tab. \ref{tab:hallucination-error-by-toolalpaca}.

\begin{table*}[t!]
\centering
\resizebox{2\columnwidth}{!}{%
\begin{tabular}
{>{\arraybackslash}m{15cm}|>{\arraybackslash}m{10cm}}
\textbf{Generated Plan by \texttt{ToolAlpaca-7B}} & \textbf{Correct APIs}\\
\toprule
{\textbf{Query: I would like to book a rental car for my visit to Las Vegas on June 28th.}} &\\
$[$thought$]$ I need to suggest cars to the customer. $[$API$]$ \colorbox{pink}{SuggestCars()} & FindRentalCar()\\
$[$thought$]$ I need to confirm and create the trip. $[$API$]$ \colorbox{pink}{ConfirmTrip()} & Confirm(), CreateTrip()\\
$[$thought$]$ I need to extract and add promotional offers. $[$API$]$ \colorbox{pink}{ExtractPromotionalOffers()} & GetCarInsuranceDiscount(), UpdateTrip()\\
$[$thought$]$ I need to order the trip. $[$API$]$ OrderTrip() & GetPaymentInformation(), OrderTrip()\\

\midrule

\toprule
{\textbf{Query: Can you help me open an investment account?}} &\\
$[$thought$]$ I need to check the eligibility of the customer to open an account. $[$API$]$ CheckEligibility() & GetCustomerIncome(), GetDateOfBirth(), CheckEligibility()\\
$[$thought$]$ Based on the eligibility, I can finalize the account type. $[$API$]$ \colorbox{pink}{FinalizeAccountType()} & GetAccountTypes()\\
$[$thought$]$ Now, I can open the account. $[$API$]$ OpenAccount() & Confirm(), OpenAccount()\\

\bottomrule
\end{tabular}}
\caption{API hallucination examples by \texttt{ToolAlpaca}. Hallucinated APIs are highlighted.}
\label{tab:hallucination-error-by-toolalpaca}
\end{table*}

\end{document}